\documentclass[journal, 10pt]{IEEEtran}
\IEEEoverridecommandlockouts
\usepackage{cite}
\usepackage{url}
\usepackage{amsmath,amssymb,amsfonts}
\usepackage{algorithmic}
\usepackage{graphicx}
\usepackage{textcomp}
\usepackage{xcolor}
\usepackage{gensymb}
\usepackage{booktabs}
\usepackage{array}
\usepackage{subcaption}
\usepackage{multirow}
\usepackage{hyperref}
\usepackage{soul}

\usepackage[normalem]{ulem}
\def\BibTeX{{\rm B\kern-.05em{\sc i\kern-.025em b}\kern-.08em
    T\kern-.1667em\lower.7ex\hbox{E}\kern-.125emX}}

\renewcommand{\baselinestretch}{0.9}
\begin{document}

\newcommand{\mm}[1]{\textcolor{green}{Michael: #1}}
\newcommand{\lo}[1]{\textcolor{orange}{Leo: #1}}
\newcommand{\rp}[1]{\textcolor{teal}{Russ: #1}}
\newcommand{\yellow}[1]{\colorbox{yellow}{#1}}
\newcommand{\inaz}[1]{\textcolor{red}{Irtija: #1}}
\newcommand{\ma}[1]{\textcolor{magenta}{#1}}
\newcommand{\fa}[1]{\textcolor{magenta}{Fatemeh: #1}}
\newcommand{\red}[1]{\textcolor{red}{ #1}}
\newcommand{\mh}[1]{\textcolor{cyan}{Mobin: #1}}

\newcommand{\ResultFigure}[4]{
\begin{figure*}[t]
  \centering
  \begin{subfigure}{0.5\textwidth}
    \centering
    \includegraphics[width=\linewidth]{#1/#2__learning_curves.png}
    \caption{Training vs. Validation}
  \end{subfigure}
  \hfill
  \begin{subfigure}{0.2\textwidth}
    \centering
    \includegraphics[width=\linewidth]{#1/#2__confusion_matrix.png}
    \caption{Confusion Matrix}
  \end{subfigure}
  \hfill
  \begin{subfigure}{0.2\textwidth}
    \centering
    \includegraphics[width=\linewidth]{#1/#2__roc_curve.png}
    \caption{ROC Curve}
  \end{subfigure}
  \caption{#3}
  \label{fig:#4}
\end{figure*}
}

\title{FLAME 3 Dataset: Unleashing the Power of Radiometric Thermal UAV Imagery for Wildfire Management\\
\thanks{This material is based upon work supported by the National Aeronautics and Space Administration (NASA) under award number 80NSSC23K1393, the National Science Foundation under Grant Numbers CNS-2232048, CNS-2038759, CNS-2038589, and CNS-2204445, Salt River Project (Award \#8200007407, and DoD SERDP Closing Gaps Project RC20-1025.)}
}

\author{
    Bryce Hopkins$^{1}$,
    Leo O'Neill$^{2}$,
    Michael Marinaccio$^{1}$,
    Mobin Habibpour$^{1}$,
     Eric Rowell$^{3}$,\\
      Russell Parsons$^{4}$,
       Sarah Flanary$^{4}$,
       Irtija Nazim$^{5}$,
       Carl Seielstad$^{6}$,
       Fatemeh Afghah$^{1}$

\thanks{$^{1}$Holcombe Department of Electrical and Computer Engineering, Clemson University, Clemson, SC, USA,
        {\tt\small \{bryceh,mmarina,fafghah\}@clemson.edu}}
\thanks{$^{2}$Pacific Southwest Research Station, U.S. Forest Services, Redding, CA  USA, 
        {\tt\small Christopher.Oneill@usda.gov}}%
\thanks{$^{3}$School of Environmental and Forest Sciences, University of Washington, Seattle, WA USA, 
        {\tt\small erowel@uw.edu}}%
\thanks{$^{4}$US Forest Service, Rocky Mountain Research Station, Fire Sciences Laboratory, Missoula, MT USA, 
        {\tt\small \{russell.a.parsons,sarah.j.flanary\}@usda.gov}}
        \thanks{$^{5}$Department of Mechanical Engineering, Clemson University, Clemson, SC, USA,
        {\tt\small inazim@clemson.edu}}
\thanks{$^{6}$ Department of Forest Management, University of Montana, Missoula, MT USA
        {\tt\small carl.seielstad@umontana.edu}}%
}

\maketitle
\begin{abstract}
The increasing accessibility of radiometric thermal imaging sensors for unmanned aerial vehicles (UAVs) offers significant potential for advancing AI-driven aerial wildfire monitoring and decision support. Radiometric imaging provides per-pixel temperature estimates, offering a quantitative alternative to non-radiometric thermal imagery where temperature information is typically visualized through vendor-specific color palettes. Despite its advantages, radiometric UAV wildfire imagery remains underutilized in the research community, in part due to the limited availability of publicly accessible datasets containing synchronized visual-spectrum and radiometric thermal data. This study addresses this gap by introducing procedures for collecting and processing synchronized visual-spectrum and radiometric long-wave infrared imagery from UAVs during prescribed fires. We further present the FLAME~3 dataset, a collection of paired visual-spectrum and radiometric thermal imagery that builds upon prior FLAME datasets by including radiometric thermal Tag Image File Format (TIFF) files and nadir thermal plot data products \cite{FLAME3Sycan,FLAME3Nadir,FLAME3-dataport}. The dataset and associated processing tools are intended to support the development and evaluation of machine learning methods for aerial wildfire detection, segmentation, and related assessment tasks.
\end{abstract}

\begin{IEEEkeywords}
wildfire, UAV, fire image dataset, radiometric thermal imagery, computer vision.

\end{IEEEkeywords}

\section{Introduction}

Monitoring wildfire for both intentional and unintentional ignitions is a crucial data source for characterizing fire behavior and rate of spread, as well as for supporting fire effects analysis, firefighter safety, and tactical decision-making for Incident Management Teams (IMTs). Accurate and timely fire intelligence is essential for understanding fire dynamics and enabling both operational response and longer-term modeling efforts. Unmanned aerial vehicles (UAVs) are playing an increasingly important role in wildfire monitoring, driven by rapid advances in onboard sensors, autonomous systems, and machine learning approaches capable of producing actionable information from aerial imagery. UAV-based approaches for wildfire monitoring differ from traditional methods employed by the U.S. Forest Service–managed National Infrared Operations (NIROPS), which are designed to produce incident-wide intelligence products that provide a broad situational overview of large fires.

UAV operations are typically used for strategic and localized monitoring of wildfire activity (e.g., division-level assessment, fire starts, safety checks, and patrol operations), due to platform endurance constraints (\raisebox{-.5ex}{\textasciitilde}25--45~min flight durations), flight altitude limitations, and airspace deconfliction requirements. The Interagency UAS Program\footnote{https://uas.nifc.gov/} is tasked with supporting wildfire response through the development of standards for UAV data collection, processing, and management in support of IMTs and broader resource management objectives. These operational standards, together with established wildfire aviation and infrared mapping practices, directly influence the types of data products that can be collected and the conditions under which UAV-based sensing is feasible. As a result, current UAV wildfire datasets are often constrained by operational priorities rather than research-driven data quality requirements.

Advantages of UAV-based sensing include rapid deployment, high spatial and temporal resolution, loitering capability, lower operational cost, and reduced risk to personnel. A key emerging area of interest in UAV wildfire operations is the development of image processing pipelines capable of rapidly transforming raw aerial imagery into usable fire intelligence products. These products may include live video streams, infrared image mosaics, vectorized fire perimeters, or dual-band visible/thermal imagery collected synchronously. Such data products have the potential to improve next-generation fire behavior models, support structure-from-motion (SfM) reconstruction of fuels and terrain, enable temporally rich observations of fire progression, and provide labeled training data for machine learning approaches targeting wildfire detection, spread estimation, and ignition monitoring. Furthermore, the provision of high-fidelity, synchronized radiometric data opens novel pathways for training multimodal large language models (MLLMs) and visual question answering (VQA) frameworks tailored to aerial fire intelligence, such as the recently proposed WildfireVQA benchmark \cite{habibpour2026wildfirevqalargescaleradiometricthermal}.

However, a critical gap remains in the availability of high-fidelity radiometric data for the research community. While recent advances in AI-based remote sensing have demonstrated strong performance for fire-related tasks, reproducible evaluation of such models under realistic wildfire conditions remains challenging. As detailed further in Section II, existing public datasets primarily consist of web-sourced imagery \cite{AIDER1, AIDER2, DINCER, MendeleyFireDataset}, ground-based closed-circuit television (CCTV) captures \cite{FireDetectionFromCCTV, Mivia1}, or non-radiometric thermal data \cite{RGB-T-China, flame_1_dataset}. Datasets suitable for studying fire behavior, thermal dynamics, and multimodal data fusion at fine spatial and temporal scales are severely limited. Consequently, much of the existing research relies on color-mapped visual representations rather than absolute temperature measurements, restricting the types of analyses and physical learning objectives that can be reliably pursued.


This work is motivated by the hypothesis that providing raw radiometric temperature data, rather than processed color-mapped images, will enable machine learning models to learn robust, physically grounded features that generalize better across varying environmental conditions. We posit that the inclusion of per-pixel temperature data allows for more precise discrimination between active fire and distinct background thermal anomalies, overcoming limitations inherent in standard RGB or non-radiometric thermal datasets.


The primary scientific contributions of this work are as follows:
\begin{itemize}
    \item We identify and characterize the critical data collection and processing challenges that currently hinder the advancement of aerial dual-band wildfire sensing, specifically regarding sensor synchronization and radiometric calibration.
    \item We propose a standardized methodological framework for maximizing the acquisition of high-fidelity aerial imagery during prescribed fires, including a semi-automated software suite for post-processing and alignment.
    \item We introduce the FLAME~3 dataset, comprising distinct Computer Vision and Modeling subsets, which serves as a new benchmark for evaluating classification, segmentation, and fire behavior modeling tasks using radiometric data.
    \item We provide a comprehensive empirical evaluation comparing the FLAME family of datasets across varying input modalities, demonstrating the superior performance and generalization capabilities enabled by the inclusion of radiometric thermal data.
\end{itemize}

The remainder of this paper is structured as follows: Section \ref{sec:related_work} reviews related work and outlines the operational and environmental challenges inherent to aerial wildland fire sensing, highlighting gaps in existing public datasets. Section \ref{sec:flame} introduces the FLAME~3 dataset, detailing the characteristics of the radiometric thermal TIFF files, image pair alignment considerations, and the nadir thermal plots. Section \ref{sec:methodology} details the rigorous methodology behind the FLAME~3 data collection and the subsequent semi-automated preprocessing and alignment pipeline. Section \ref{sec:classification_results} presents our empirical classification results, including data starvation ablation studies and multimodal fusion comparisons, to validate the utility of the radiometric dataset. Finally, Section \ref{sec:conclusions} concludes the paper with a discussion on dataset limitations and broader implications for the field.

\section{Related Work and Wildfire Data Collection Challenges}
\label{sec:related_work}

Wildfire, defined as a rapidly spreading and uncontrolled fire event, presents a complex and hazardous environment for UAV-based data collection, where fire suppression, responder safety, and operational priorities take precedence over research-oriented data acquisition. As a result, research access to active wildfire environments is limited by UAV platform restrictions, agency-specific standards and certifications, and incident management team (IMT) resource ordering requirements. To date, most UAV-based wildfire data collection efforts have been conducted directly by federal or state agency UAS programs or through approved private contractors operating under strict operational protocols\cite{NWCG_Standards}.

Research-oriented UAV data collection has therefore largely focused on prescribed fires, which are planned, lower-intensity burns conducted under controlled environmental and operational conditions. Prescribed fires offer greater accessibility for researchers due to more flexible airspace management, predictable burn plans, and broader land ownership participation (e.g., private, state, and federal lands). However, reliance on prescribed fire data introduces limitations in the range of fire behavior captured, as these burns typically occur under higher fuel moisture conditions and are designed to limit extreme fire behavior. Consequently, prescribed-fire imagery may not fully represent the intensity, plume dynamics, or rapid spread observed in large, uncontrolled wildfire incidents \cite{BOROUJENI2024102369}.

\subsection{Wildfire Datasets}
\begingroup 
\begin{table*}[htbp]
\caption{Existing Fire/Wildfire Imagery Datasets vs. FLAME Datasets}
\label{Table:RelatedDatasets}
\setlength{\tabcolsep}{4pt} 
\renewcommand{\arraystretch}{1.6} 
\begin{center}
\begin{tabular}{
    >{\centering\arraybackslash}m{0.20\linewidth} 
    >{\centering\arraybackslash}m{0.13\linewidth}
    >{\centering\arraybackslash}m{0.12\linewidth} 
    >{\centering\arraybackslash}m{0.10\linewidth}
    >{\centering\arraybackslash}m{0.08\linewidth}
    >{\centering\arraybackslash}m{0.09\linewidth} 
    >{\centering\arraybackslash}m{0.08\linewidth}
    >{\centering\arraybackslash}m{0.06\linewidth} 
}
\toprule
\textbf{Dataset Name/Year} & \textbf{Collection, Perspective} & \textbf{Image Type} & \textbf{Landscape} & \textbf{Pre/Post Burn Data} & \textbf{Radiometric Data} & \textbf{Applications} & \textbf{Image Count} \\[0.2mm]

\midrule
        UAVs-FFDB, $2024^a$\cite{ffdb2024} & UAV mounted RaspiCamV2, Aerial & RGB & Rural Pile Burns & Pre-burn Imagery Only & No & Classification, Modeling, and Segmentation & 1653$^d$ \\
        
        State Key Laboratory of Fire Science, University of Science and Technology of China Dataset, 2023 $^a$ \cite{RUI2023103554, RGB-T-China} & RGB Camera and UAV, Ground and Aerial & RGB/Thermal Pairs & Urban & No & No & Semantic Segmentation & 1,367$^b$ \\
        
        DataCluster Labs' Fire and Smoke Dataset, 2021 \cite{DataClusterLabs} & Cellular Camera, Ground & RGB & Mixed$^a$ & No & No & Classification & 7000+$^c$ \\

        Dincer Wildfire Detection Image Data, 2021 \cite{DINCER} & Web Sourcing, Ground &  RGB & Rural & No & No & Classification & 1,900$^d$ \\

        AIDER, 2020 \cite{AIDER1, AIDER2} & Web Sourcing, Aerial & RGB & Assorted & No & No & Classification & 1,000$^d$ \\

        Dataset for Forest Fire Detection, 2020 \cite{MendeleyFireDataset} & Web Sourcing, Aerial and Ground & RGB & Rural & No & No & Classification & 1,900$^d$ \\

        Fire Detection by Dhruvil Shah, 2020 \cite{jackfrost} & Web Sourcing, Ground & RGB & Mainly Urban & No & No & Classification & 3,225$^d$ \\
        
        FireNet, 2019 \cite{FireNET} & Web Sourcing, Ground & RGB & Assorted & No & No & Object Detection & 502$^d$ \\

        Fire Detection From closed-circuit television (CCTV), 2019 \cite{FireDetectionFromCCTV} & CCTV, Ground & RGB & Assorted & No & No & Classification & 864$^c$ \\

        Furg Fire Dataset, 2018 \cite{FURG1, FURG2} & Web Sourcing, Ground & RGB & Urban & No & No & Object Detection & 365,702$^c$ \\

        CAIR's Fire Detection Image Dataset, 2017\cite{CAIR} & Web Sourcing, Ground & RGB & Mainly Urban & No & No & Classification & 651$^d$ \\

        Corsican Fire Dataset, 2017 \cite{TOULOUSE2017188} & RGB Camera, Ground & RGB/Thermal Pairs & Mainly Urban & No & No & Semantic Segmentation & 635$^b$ \\

        Mivia's Fire Detection Dataset, 2014 \cite{Mivia1, Mivia2} & CCTV, Ground & RGB & Assorted & No & No & Classification & 62,690$^c$ \\
\midrule
        FLAME, 2020 $^a$$^f$\cite{shamsoshoara2021aerial} & UAV, Aerial & RGB/Thermal, Not Pairs & Rural Pile Burns & No & No & Classification and Segmentation & 47,992$^c$ \\

        FLAME 2, 2022 $^a$$^g$ \cite{FLAME2Dataset} & UAV, Aerial & RGB/Thermal Pairs & Rural Prescribed Burns & Yes$^j$ & No & Classification & 53,451$^b$$^c$ \\

        \textbf{FLAME 3}, 2024 $^a$$^h$ & UAV, Aerial & side-by-side Dual RGB/IR and Thermal TIFF & Rural Prescribed Burns & Yes$^k$ & Yes & Classification, Modeling, and Segmentation & 13,997$^e$\\

        \multicolumn{1}{c|}{} & \multicolumn{7}{l}{\textbf{F3 CV Subset (IR, RGB)  - Sycan Marsh: (738, 738); Wilamette: (233, 233); Shoetank: (3540, 3540)}} \\

        \multicolumn{1}{c|}{} & \multicolumn{7}{l}{\textbf{F3 Modeling Subset (IR, RGB) - Hanna Hammock, (800, 400), Nadir Thermal Plots}} \\
        
\bottomrule
\end{tabular}
\end{center}
\footnotesize
\vspace{2mm}
$^a$ Contains supplemental data, examples include, but are not limited to: weather data, 3D point cloud data, augmented imagery, and thermal TIFF files \\
$^b$ Refers to RGB/IR image pairs, both showing the same scene \\
$^c$ Refers to extracted video frames, rather than independently collected raw images \\
$^d$ Refers to raw images, not image pairs or video frames \\
$^e$ Total images, some paired RGB/IR, some not \\
$^f$ \textbf{FLAME 1} extra data data includes three types of thermal imagery (Fusion, WhiteHot, and GreenHot), as well as a 2,003 subset of segmentation masks\\
$^g$ \textbf{FLAME 2}, aerial multispectral dataset, extra data includes weather data, preburn videos, burn planning information, preburn pointcloud data, and a preburn orthomosaic\\
$^h$ \textbf{FLAME 3}, aerial multispectral dataset; Extra data contains 3D pointcloud, thermal TIFF files for RGB/Thermal Pairs, pre/post burn imagery, burn plot maps, weather data; public release subsets - CV no supplemental data, Modeling contains supplemental data\\

\end{table*}
\endgroup

In recent years, numerous fire and wildfire imagery datasets have been released, targeting general fire scenes, prescribed burns, or active wildfires depending on collection methodology and application focus. While wildfire-specific datasets are the most representative of operational conditions, their scarcity remains a significant challenge. Prescribed-fire and general fire datasets, although limited in behavioral diversity, retain value due to shared physical characteristics with wildfires and are commonly leveraged for pretraining or transfer learning scenarios. Existing datasets support a wide range of research tasks, including classification, segmentation, three-dimensional fire reconstruction, and fire spread modeling \cite{ShamsoshoaraDataset}.

Fire imagery datasets vary widely in sensing platform, modality, spatial resolution, and annotation strategy. Data sources include handheld or fixed ground cameras \cite{RGB-T-China,DataClusterLabs,TOULOUSE2017188}, satellite imagery \cite{CoenSchroed2013}, and UAV-based collections \cite{shamsoshoara2021aerial,FLAME2Dataset}. Most publicly available datasets focus on image-level classification tasks \cite{DataClusterLabs,AIDER2,FireDetectionFromCCTV,FLAME2Dataset}, while fewer provide pixel-wise annotations for segmentation \cite{RGB-T-China,TOULOUSE2017188} or bounding-box localization \cite{FURG1,FURG2}. To the best of our knowledge, representative fire imagery datasets and their characteristics are summarized in Table~\ref{Table:RelatedDatasets}.

\subsection{Gaps in Existing Datasets and Related Concerns}
\label{sec:dataset_gaps}

While wildfire imagery datasets have improved in availability and diversity, notable gaps remain that hinder progress in tasks such as detection, monitoring, modeling, post-fire assessment, and prescribed fire planning. Table~\ref{Table:RelatedDatasets} highlights these gaps by contrasting existing datasets with FLAME~3. In the following subsections, we outline key challenges related to spatial resolution, temporal resolution, and multi-sensor data collection that motivate the design of the FLAME~3 dataset.

\subsubsection{Spatial and Temporal Resolution}

A persistent challenge in aerial wildfire imagery datasets is insufficient spatial resolution, where individual image pixels correspond to large ground areas and fail to capture fine-scale fire features. In the wildfire domain, reduced spatial resolution limits the effectiveness of detection and assessment models and impairs generalization across flight altitudes. Thermal imagery is particularly susceptible to reduced spatial resolution, as long-wave infrared (LWIR) sensors typically employ smaller detector arrays and optical systems than visible-spectrum cameras, resulting in coarser ground sampling distance (GSD) at equivalent flight altitudes \cite{Schott2007RemoteSensing}. Camera resolution, defined by the number of image pixels captured (e.g., 4000$\times$3000 for RGB versus 640$\times$512 for many thermal sensors), directly influences the achievable spatial resolution on the ground.

Web-sourced imagery further exacerbates this issue, as image quality and resolution are inconsistent and often unknown \cite{AIDER1,AIDER2,DINCER}. Targeted data collections may also rely on lower-resolution thermal sensors, limiting downstream applicability \cite{ffdb2024,RGB-T-China}. Prior work has explored techniques such as interpolation from vibrating IR platforms to enhance effective thermal resolution \cite{van1999}, but such approaches are not widely adopted in publicly available datasets.

One of the primary contributions of FLAME~3 is the provision of data and collection procedures designed to achieve high spatial resolution from UAV platforms operating within regulatory constraints. FLAME~3 includes high-resolution RGB imagery (up to 4000$\times$3000 pixels) captured by DJI M30T and M2EA platforms at altitudes up to the FAA-regulated maximum of 122~m AGL \cite{FLAME3Sycan}. At this altitude, nadir RGB–thermal image pairs collected by a DJI M30T yield approximate ground resolutions of $\sim$4.4~cm/px for RGB imagery and $\sim$17.6~cm/px for thermal imagery, enabling fine-grained observation of fire fronts and thermal gradients.

Temporal resolution, defined as the revisit rate at which the same geographic area is imaged, is equally critical for wildfire monitoring and modeling. Tasks such as fire spread estimation and burn behavior characterization require repeated observations over time. However, the majority of existing datasets are web-sourced and lack temporal continuity, rendering them unsuitable for such analyses \cite{AIDER1,AIDER2,DINCER,CAIR,FireNET}. In this work, we consider temporal resolution to be sufficient when the sampling interval resolves observable displacement of the flaming front over time, allowing changes in fire position and intensity to be tracked at centimeter-to-meter scales depending on georeferencing accuracy and rate of spread. Given typical prescribed fire rates of spread (ranging from 0.1 to 1.0 m/min in surface fuels), the 1-minute revisit interval provided by our flight procedures ensures sufficient overlap to track front progression while minimizing data redundancy.

\begin{figure}[h]
  \centering
  \includegraphics[width=0.45\textwidth]{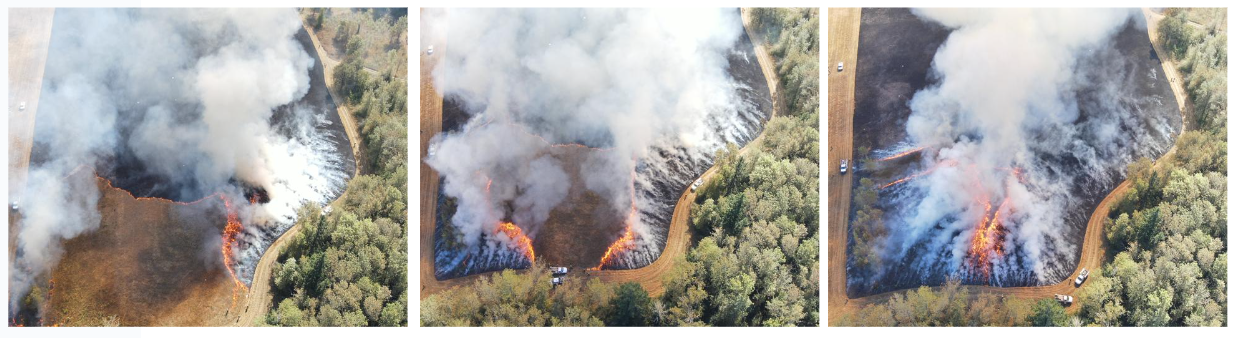}
  \caption{Three Images from Willamette prescribed burn on September 23rd, 2022. From left to right, images taken at 2:38 PM, 2:39 PM, and 2:40 PM respectively. Images were acquired during a continuous loitering flight pattern, enabling $\sim$1-minute revisit intervals of the same geographic area. Crops are aligned to show the same geographic extent.}
  \label{fig:TemporalFigure}
  \vspace{-5pt}
\end{figure}

To address this need, FLAME~3 introduces \emph{nadir thermal plots}, which provide repeated, georeferenced thermal observations of a fixed area over the duration of a burn \cite{FLAME3Nadir}. Figure~\ref{fig:TemporalFigure} illustrates an example of repeated sampling in the visible spectrum, capturing progressive burn development. These images were acquired during a single UAV flight using loitering and overlapping flight paths to revisit the same area at approximately one-minute intervals. While high temporal resolution enables detailed fire progression analysis, it may reduce data variability for tasks such as classification. FLAME~3 therefore balances high-temporal-resolution nadir plots with lower-temporal-resolution oblique imagery to support a range of downstream applications.

\subsubsection{Multi-sensor Data Collection}

Conventional computer vision approaches rely primarily on visible-spectrum imagery and have demonstrated strong performance for wildfire detection and assessment under favorable conditions. However, visual-spectrum-only models are inherently limited by their sensitivity to smoke, cloud cover, and illumination variability. Augmenting RGB imagery with infrared sensing provides complementary physical information that is less affected by optical obscurants and directly correlates with fire presence.

This work employs dual-band visible-spectrum and long-wave infrared (LWIR) radiometric imaging, enabling models to leverage both visual context and temperature-based cues. Radiometric thermal imagery provides per-pixel temperature estimates at moderate resolutions (e.g., 640$\times$512), allowing fires to be detected even when flames are partially or fully obscured by smoke. Prior research has demonstrated the benefits of RGB–IR fusion in related domains, such as urban scene segmentation using architectures like DooDLeNet \cite{doodlenet} and FEANet \cite{FEANet}. However, such approaches remain underexplored in UAV-based wildfire sensing due to the lack of suitable datasets.

A key barrier to broader adoption of multi-sensor approaches is the scarcity of high-quality paired RGB–thermal wildfire imagery. Web-sourced imagery rarely includes synchronized dual-band data, and existing UAV-based datasets often suffer from limited quantity, reduced variability, or alignment inconsistencies \cite{RGB-T-China,flame_1_dataset,TOULOUSE2017188}. FLAME~3 addresses these challenges through targeted data collection and preprocessing workflows that aim to improve alignment between visible and thermal imagery while preserving radiometric fidelity.

\subsubsection{Variety and Diversity of Data}
For datasets generated from individual targeted data collections, it is common for each sample to be captured in the same environment under the same conditions, resulting in a high level of similarity between collected samples. For example, the FLAME 1 dataset \cite{flame_1_dataset} was collected from a single prescribed fire and includes a resulting limited data diversity. Many other targeted data collection style datasets suffer from similar sample similarity \cite{RGB-T-China, TOULOUSE2017188}. As such, models trained on these datasets tend to suffer when generalizing to unseen fuel types, terrain, geographic location, weather, and/or time of day.  FLAME 3 addresses this concern by providing aerial imagery from six prescribed fire environments, encompassing a variety of fuel types, burn behaviors, collection conditions, and geographic locations. As a result, it has been found that models trained on FLAME 3 show improved generalization as compared to the previous FLAME 1 or FLAME 2 datasets, as demonstrated in Section~\ref{sec:classification_results}.

\subsubsection{Monitoring, Prescribed Fire Planning, and Post-Assessment}

In recent years, wildfire monitoring research has progressed, and many works utilizing the FLAME 1 \cite{flame_1_dataset} and FLAME 2 \cite{flame_2_dataset} datasets have been published. IC-GAN \cite{pedram_gan} was introduced for RGB-to-IR wildfire image translation. AttentionGAN \cite{attention_gan} introduces an attention-guided network focusing on image-to-image translation to create synthetic wildfire images. FLAME Diffuser \cite{pedram_flamediffuser} utilizes a variational autoencoder and diffusion models to generate synthetic, ground-truth paired wildfire imagery. FLAME Diffuser also uses the CLIP network \cite{CLIP} for text-to-image background control. The authors in  \cite{hossein_thermal} introduce an approach for generating a large wildfire dataset, utilizing a dual-GAN approach based on CycleGAN \cite{cyclegan}.  FlameFinder, an attention-based network for improving flame detection accuracy on images where wildfire is blocked by smoke was proposed in \cite{hossein_flame}. Hardware acceleration techniques  have been explored aiming towards real-time wildfire detection \cite{austin_hardware}.

While these approaches have advanced wildfire detection, complex tasks like predicting wildfire severity, modeling its spread, and analyzing fire behavior remain challenging due to limited availability of supporting datasets \cite{9217742,9049048}. All datasets listed in {Table \ref{Table:RelatedDatasets}} lack supplementary data required for modeling purposes, except for FLAME 2 \cite{FLAME2Dataset} and FLAME 3. Fuel characteristics (e.g. fuel type) moisture content, area vegetation, etc. are essential, as evident through analysis  in \cite{alizadeh2024land}. Topographic data is important to understand the variation in the terrain, which can help model how the fire may spread. Additionally, utilizing past wildfire data from the region of future UAV deployment could benefit a model in training, providing it with a better understanding of how fires behave in that environment. FLAME 3 provides a comprehensive set of supplementary data including 3D pointcloud data, thermal TIFF files, pre/post burn imagery, burn plot maps, collection procedures, and weather data to address these concerns.

Furthermore, the majority of these datasets lack information or provide only limited details on burn severity following a wildfire. Post-assessment data is crucial, as it offers insights into the affected environment, the damages incurred, and potential causes of future fires \cite{MODISPostBurn,alizadeh2024land}. Information on terrain changes resulting from wildfire damage is often missing, which hinders the assessment of both short- and long-term impacts. For tasks that rely on post-burn analysis, this data is essential. Therefore, invested in collecting supplemental, modeling-focused data from prescribed fires, as outlined in this work. Such efforts not only enhance the value of individual fire events for the broader research community but also maximize the data gathered from each occurrence.

\section{Flame~3 Dataset}

\label{sec:flame}
\begin{figure*}[t]
  \centering
  \includegraphics[width=0.9\textwidth]{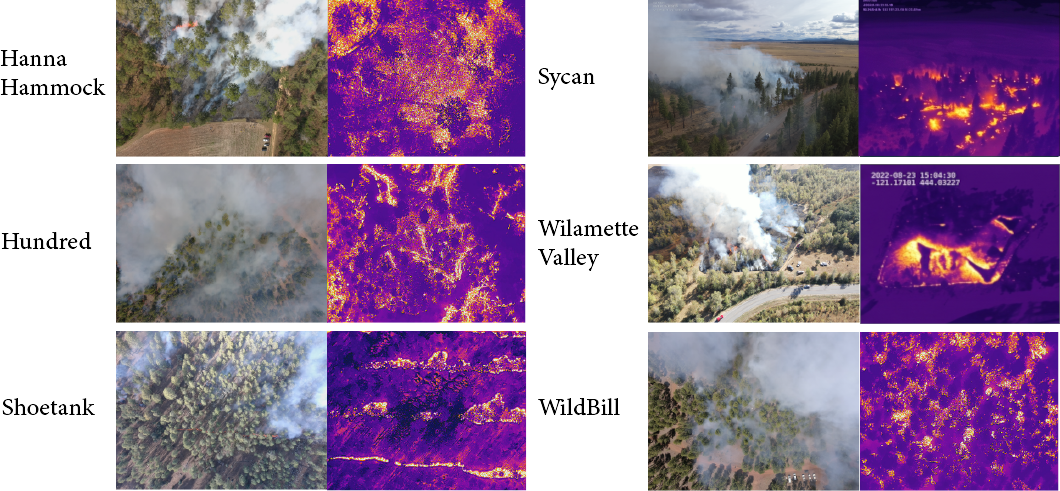}
  \caption{Raw RGB--thermal image pairs collected from each prescribed burn site listed in Table~\ref{rx_fires}. The thermal imagery utilizes a radiometric colormap where darker regions (e.g., black/dark purple) indicate cool ambient backgrounds (typically $0^\circ$C to $25^\circ$C), and brighter regions (transitioning through orange and yellow to white) represent intense heat from active combustion and heavy fuel loading (up to $650^\circ$C).}
  \label{fig:FireBurnFigure}
  \vspace{-5pt}
\end{figure*}

The Flame 3 dataset consists of paired visual spectrum and long-wave infrared imagery from various prescribed burns \cite{FLAME3Sycan}. This dataset includes generalized aerial oblique imagery collected via UAS at six different prescribed fire events, which showcase diverse burn behaviors across pine forests, grasslands, and low sagebrush environments. Additionally, nadir thermal plots were established at three of these six burns, resulting in a focused subset of FLAME 3 that facilitates the characterization of fine-grained temporal and spatial dynamics and supports predictive modeling \cite{FLAME3Nadir}. Table \ref{rx_fires} provides a detailed overview of each of the six burns, and Figure~\ref{fig:FireBurnFigure} displays raw, unprocessed sample images from each burn site.

\subsection{Radiometric Thermal TIFF}

\subsubsection{Pixel-level Fire/No-Fire Labeling and Temperature Ground Truth}

A critical step in supervised learning for wildfire monitoring tasks is the generation of reliable labels aligned with the intended learning objective. Traditional annotation pipelines rely heavily on manual labeling, where human annotators assign image-level or pixel-level labels through graphical tools or text-based metadata. Such approaches are time-consuming, labor-intensive, and prone to subjectivity, particularly for large-scale datasets and ambiguous fire scenes.

Radiometric thermal TIFF files provide a distinct advantage by embedding quantitative temperature measurements directly at the pixel level. Each thermal TIFF included in FLAME~3 is a two-dimensional raster in which each pixel stores a floating-point temperature value in degrees Celsius, derived from the sensor’s calibrated radiometric response. Unlike standard 8-bit thermal images which only encode relative thermal contrast, this format preserves absolute temperature data, enabling the derivation of physical fire parameters such as fire radiative power (FRP) and intensity.

To reduce the burden of manual annotation for binary Fire/No-Fire classification, FLAME~3 leverages temperature-based heuristics derived from the radiometric TIFF data. Images with a maximum pixel temperature below approximately 80$^{\circ}$C are highly likely to correspond to background or no-fire conditions, while images exhibiting temperatures exceeding approximately 200$^{\circ}$C typically contain active flaming combustion. These thresholds were selected based on empirical analysis of our specific prescribed fire data and are consistent with wildland fire temperature ranges reported in literature \cite{Butler_Temp, Dickinson_Temp}.

It is important to note that these threshold values are not universal and may vary depending on sensor calibration, emissivity assumptions, ambient environmental conditions, and fire behavior. Accordingly, FLAME~3 provides these thresholds as practical heuristics rather than fixed rules, and users are encouraged to recalibrate or adapt them to their specific sensing platform and application context. Images that fall outside these confident threshold ranges are manually reviewed or excluded to ensure labeling reliability.

Beyond image-level classification, radiometric TIFFs also enable pixel-wise annotation for segmentation and regression-based tasks. Through direct temperature thresholding or more advanced methods such as Otsu’s method \cite{Otsu1979} and hysteresis thresholding \cite{Canny1986}, individual pixels can be algorithmically classified based on thermal intensity. Moreover, the availability of per-pixel temperature ground truth allows models to be trained for temperature estimation or thermal regression using visible-spectrum or non-radiometric thermal imagery as input, with radiometric TIFFs serving as the reference signal.

The effectiveness of pixel-level labeling and regression tasks depends critically on spatial alignment between radiometric thermal data and corresponding visible-spectrum imagery. As discussed in the following subsection, residual misalignment between RGB and thermal sensors remains a non-trivial challenge and represents an important limitation for fine-grained multimodal fusion.

\subsubsection{Image Pair Alignment}

For FLAME~3 data collection, UAV platforms equipped with side-by-side visible-spectrum and thermal cameras were employed, resulting in synchronized but not perfectly aligned image pairs. Differences in focal length, sensor resolution, lens distortion, and physical camera placement lead to systematic field-of-view (FOV) mismatches between RGB and thermal images, even when timestamps are synchronized. Such discrepancies complicate pixel-wise correspondence and can introduce errors in multimodal learning tasks.

To mitigate these effects, FLAME~3 includes a preprocessing workflow that applies manual FOV correction to visible-spectrum imagery to better match the spatial extent and resolution of the thermal images. RGB images are cropped and scaled to approximate one-to-one correspondence with thermal imagery, reducing gross spatial misalignment while preserving radiometric integrity of the thermal data.

Despite these corrections, residual misalignment remains due to lens distortion and perspective differences, particularly toward image edges. Alignment error was evaluated by overlaying salient thermal features onto RGB imagery and measuring pixel-wise displacement using Euclidean distance, revealing offsets of up to approximately 20 pixels in certain regions (Figure~\ref{fig:AlignmentErrorFigure}). While sufficient for image-level classification and coarse fusion, such errors may impact fine-grained pixel-level fusion or regression tasks.

Future work will explore automated distortion correction and calibration-based alignment techniques to further reduce residual offsets and enable more precise multimodal correspondence.

\subsubsection{TIFFs vs.\ JPEGs}

Radiometric thermal cameras record infrared radiation as a single-band raster signal, where each pixel corresponds to a measured temperature or radiance value. This data is naturally stored in formats such as TIFF, which support high-precision numerical values without normalization. In contrast, commonly distributed thermal JPEG or PNG images are generated by mapping temperature values to RGB colors using proprietary or vendor-specific color palettes, primarily for human interpretability.

While color-mapped thermal images are visually intuitive, the conversion process normalizes temperature values into a limited 8-bit range (0–255), resulting in information loss and reduced numerical fidelity. This normalization can obscure subtle thermal gradients and introduces palette-dependent artifacts that are undesirable for machine learning applications relying on physically meaningful input features. As a result, models trained directly on color-mapped thermal imagery may learn spurious correlations tied to palette choice rather than underlying thermal structure.

FLAME~3 therefore provides both radiometric thermal TIFF files and corresponding color-mapped thermal JPEGs. The TIFF files preserve raw temperature measurements for machine learning, quantitative analysis, and ground-truth generation, while the JPEGs support visualization and qualitative inspection. To the authors’ knowledge, FLAME~3 is the first publicly available UAV-collected wildfire imagery dataset to include paired visible-spectrum imagery and radiometric thermal TIFFs \cite{FLAME3-dataport,Micheal_Asilomar25}. For reference, the FLIR dataset \cite{FLIR} provides radiometric thermal TIFFs for urban object detection but does not address wildfire sensing scenarios \cite{TIFFPaper}.

\subsection{Nadir Thermal Plots}
\begin{figure}[h]
  \centering
  \includegraphics[width=0.45\textwidth]{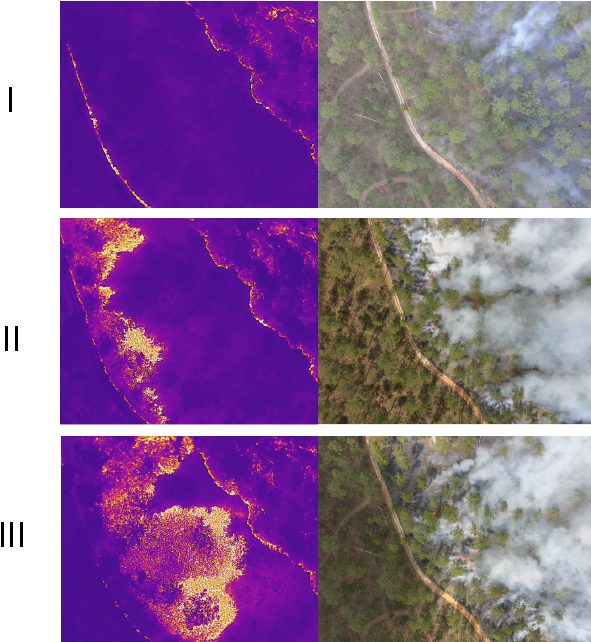}
  \caption{RGB-IR Image Pairs from Hanna Hammock prescribed burn. From top to bottom, time progresses between images with images taken at 5:48:30 PM, 5:48:42 PM, and 5:48:50 PM.}
  \label{fig:NadirPlotTemporalFigure}
  \vspace{-5pt}
\end{figure}

The FLAME 3 Nadir Thermal Plot datatype provides a persistent overhead recording of a fire's progression across a set georeferenced plot. While both visible spectrum and thermal imagery is typically captured, they are referred to as nadir \textit{thermal} plots as the radiometric thermal data is typically the most interesting and informative aspect of the data. During collection, operators locate a UAV above the center of a preset georeferenced area, configuring the UAV to take repeat imagery every 3-5 seconds. When collated, orthorectified, and appropriately stacked, the collected data encapsulates burn progression behaviors across the studied plot fuels. With appropriate calibration, the resulting data product can be used to measure centimeter-grade rate of spread as well as evaluate energy release \cite{Oneill}. Figure~\ref{fig:NadirPlotTemporalFigure} shows a few nadir thermal plot samples showing burn progression over time.

\begin{table*}[htbp]
\caption{Descriptions of prescribed fires attended. ROS is rate of spread, a measure of flaming front velocity}
\label{rx_fires}
\setlength{\tabcolsep}{4pt} 
\renewcommand{\arraystretch}{1.3} 
\begin{center}
\begin{tabular}{>
{\centering\arraybackslash}m{0.09\linewidth} >{\centering\arraybackslash}m{0.10\linewidth} >{\centering\arraybackslash}m{0.09\linewidth} >{\centering\arraybackslash}m{0.09\linewidth} >{\centering\arraybackslash}m{0.08\linewidth} >{\centering\arraybackslash}m{0.20\linewidth} >{\centering\arraybackslash}m{0.09\linewidth} >
{\centering\arraybackslash}m{0.07\linewidth} >
{\centering\arraybackslash}m{0.05\linewidth}}
\toprule
\textbf{Prescribed Fire} & \textbf{Location (elevation, m)} & \textbf{Date} & \textbf{Forest Type} & \textbf{Avg. Fuel Moisture (\%)$^a$} & \textbf{Fire behavior (ocular estimate)} & \textbf{No. of images (IR, RGB)$^b$} & \textbf{Pre- and post-fire collection} & \textbf{Nadir Plot}\\[0.2mm]
\midrule

        Hanna Hammock & Tall Timbers Research Station, Florida (60) & February 2023 & Longleaf Pine & 15.0 & surface head fire w/ rapid ROS, inactive backing fire, short residence time & 800, 400 & Y & Y\\

        Hundred & Tonto National Forest, Arizona (2,000) & April, 2023 & Ponderosa Pine & 4.2 & moderate intensity surface fire with low ROS, consumption of woody debris & 1600, 800 & Y & Y\\

        Shoetank Rx Burns Units 1 and 2 & San Carlos Apache Tribe, Arizona (1,900) & November 2nd, 2022 & Ponderosa Pine, Grass, Sagebrush & 7.0 & very low intensity creeping fire, some unburned patches & 3540, 3540 & Y & N\\

        Sycan & TNC Sycan Marsh Preserve, Oregon (1,500) & October 25-27, 2022 & Ponderosa pine, grass, low sage & 13.5 & low ROS, patchy fire spread in surface fuels but good consumption in heavy fuels. Isolated torching & 738, 738 & Post only & N\\

        Willamette Valley & Willow Creek Preserve (130-560) & September 23-24, and October 2022 & Oak woodland, upland grass & 8.6 &  low to high ROS continuous fire spread in grass fuels, varied behavior in oak woodland patches & 1819, 1819 & Y & N\\

        Wildbill & Coconino National Forest, Arizona (2,400) & May/June, 2023 & Ponderosa Pine & 5.9 & moderate ROS and intensity, long residence time while consuming woody debris, some isolated tree torching (plot 3) & 5500, 1400 & Y & Y\\

\bottomrule
\end{tabular}
\vspace{-10pt}
\end{center}

\footnotesize{\vspace{1mm}$^a$ Estimated using the average daily 10-hour fuel moisture from the nearest Remote Automated Weather Station, available from the Western Regional Climate Center (http://www.raws.dri.edu).} \\
\footnotesize{$^b$ A larger number of IR images indicates that a nadir thermal plot was set-up at the fire. This Includes Hanna Hammock, Hundred, and Wild Bill.}
\end{table*}

\begin{table*}[htbp]
\caption{Data Collection Equipment / Camera Settings by Prescribed Fire}
\label{equipment}
\setlength{\tabcolsep}{4pt} 
\renewcommand{\arraystretch}{1.3} 
\begin{center}
\resizebox{\textwidth}{!}{
\begin{tabular}{>{\centering\arraybackslash}m{0.15\linewidth} >{\centering\arraybackslash}m{0.20\linewidth} >{\centering\arraybackslash}m{0.10\linewidth} >{\centering\arraybackslash}m{0.20\linewidth} >{\centering\arraybackslash}m{0.20\linewidth} >
{\centering\arraybackslash}m{0.20\linewidth}
}
\toprule
\textbf{Prescribed Fire} & \textbf{Drone Used} & \textbf{Camera Model} & \textbf{RGB Resolution (WxH)$^a$} & \textbf{IR Resolution$^a$} & \textbf{Saturation Range (Low Gain; High Gain)}\\[0.2mm]
\midrule
Hanna Hammock $^b$ $^c$ & Autel EVO II Dual 640T Enterprise & XT709 & 8000x6000 & 640x512 & 0° to 550°C; -20° to 150°C \\
Hundred $^b$ $^c$ & Autel EVO II Dual 640T Enterprise & XT709 & 4000x3000 & 640x512 & 0° to 550°C; -20° to 150°C\\
Shoetank Rx Burns Units 1 and 2 $^b$ $^c$& DJI Mavic 2 Enterprise Advanced & M2EA & 8000x6000 & 640x512 & -20° to 450°C; -20° to 150°C\\
Sycan $^b$ $^c$& DJI Matrice 30T & M30T & 4000x3000 & 640x512 & 0° to 500°C; -20° to 150°C\\
Willamette Valley $^b$ $^c$& DJI Mavic 2 Enterprise Advanced & M2EA & 8000x6000 & 640x512 & -20° to 450°C; -20° to 150°C\\
Wildbill $^b$ $^c$& Autel EVO II Dual 640T Enterprise & XT709 & 7680x4320 & 640x512 & 0° to 550°C; -20° to 150°C\\
\bottomrule
\end{tabular}}
\vspace{-10pt}
\end{center}

\footnotesize{\vspace{1mm}$^a$ Refers to raw resolution that image was captured at during data collection, not necessarily camera's native resolution} \\
\footnotesize{$^b$Additional Technical Equipment: RTK GPS (1) for georeferencing point clouds, reflective aluminum plates (4) for georeferencing and marking plots} \\
\footnotesize{$^c$Supplemental Equipment: weather station, Willtronics fuel moisture meter (or other method to collect, dry, and weight fuel for percent moisture), plot frame, paper or burlap bags, clippers} \\ 
\footnotesize{At the standard 122 m AGL flight altitude, these resolutions yield an approximate Ground Sampling Distance (GSD) of ~4.4 cm/px for RGB and ~17.6 cm/px for Thermal imagery}
\end{table*}

\section{Methodology} 
\label{sec:methodology}

To encourage and enhance UAS-collected wildfire imagery, this section describes the procedures and rationale used to collect image data at the prescribed burns included in FLAME~3. We also provide two python programs to assist with post-collection data processing, focusing specifically on unlocking radiometric thermal data’s underutilized capabilities for ground truthing in machine learning applications. Finally, we provide a single-burn computer-vision-oriented subset of example post-processed imagery data collected at a prescribed fire in the pine forests at Sycan Marsh, Oregon in October 2022 \cite{FLAME3Sycan}. We also provide a single-burn modeling-focused subset including nadir thermal plot data from a prescribed fire in the tree stands at Hanna Hammock, Florida in February 2023 \cite{FLAME3Nadir}.


\subsection{UAV Data Collection}

Aerial data were collected using DJI M30T and M2EA UAV platforms equipped with integrated visible-spectrum and long-wave infrared (LWIR) thermal sensors. Flights were conducted during prescribed fire operations under approved airspace authorizations and in coordination with burn managers to ensure operational safety.

RGB imagery was captured at a native resolution of up to 4000$\times$3000 pixels, while thermal imagery was captured at 640$\times$512 pixels in radiometric mode. All imagery was time-synchronized and geotagged using onboard GPS and inertial measurements. Data were collected from altitudes ranging from approximately 61-122~m above ground level, subject to regulatory constraints and fire behavior.

\subsection{Data Collection Procedures}
With Federal Aviation Administration (FAA) and United States Forest Service (USFS) operational restrictions, short weather-based notices, remote burn locations, and limited burn seasons, attending prescribed fires to collect data is challenging. Thus, it is in a researcher’s best interest to collect as much data per burn as possible. Data collection itself is complicated by uncertain flight restrictions during the burn that are necessary to avoid interfering with burn management operations. As such, it is important to incorporate flexibility into data collection procedures. The following subsections detail data collection goals for FLAME~3 prescribed fires across pre-burn, active-burn, and post-burn phases. Imagery was collected to support both machine learning and wildfire modeling applications. All data collection equipment by prescribed burn location is organized in \textbf{Table \ref{equipment}}.

Thermal cameras operate within a specified saturation range, which defines the temperature interval over which radiometric measurements maintain accuracy. Values outside of this range result in sensor saturation, causing recorded temperatures to become clipped and unreliable. Table \ref{equipment} reports the saturation ranges for both low-gain and high-gain modes for each thermal camera used in FLAME~3 data collection. For FLAME~3 data collection, low-gain mode was used to maximize measurable temperature range and ensure accurate capture of prescribed fire temperatures. Natural wildfire flame temperatures typically range between $800^{\circ}$C and $1000^{\circ}$C, but can vary depending on burn severity. FLAME~3 data collection was conducted exclusively at prescribed fires, which burn at substantially lower temperatures and reduced intensity compared to naturally occurring wildfires. Expected temperature ranges therefore varied across FLAME~3 locations due to differences in fuel type, fuel moisture, and burn conditions. Typical prescribed fire temperatures observed ranged from 55$^{\circ}$C--~650$^{\circ}$C.

Depending on the imaging system used, radiometric calibration procedures differed. For DJI-branded sensors used in this work, radiometric recalibration is performed annually during manufacturer servicing, enabling direct conversion of sensor emission counts to temperature values in degrees Celsius using internally stored calibration curves. For the Autel UAS employed in this work, blackbody-based radiometric calibration was performed manually. While the full calibration procedure is outside the scope of this paper, the complete methodology used to calibrate and convert Autel thermal measurements to temperature values is detailed in \cite{Oneill}.

\subsubsection{Pre-Burn}
The FLAME~3 pre-burn data collection objectives were as follows:

\begin{itemize}
    \item Pre-burn structure-from-motion (SfM) photogrammetry
    \item Generic No-Fire oblique imagery
    \item Place and georeference metal plates for nadir thermal plot(s)
    \item (Supplemental) Setup weather station
\end{itemize}

Structure-from-motion (SfM) photogrammetry was performed over the burn plot prior to ignition. This step provides a baseline digital elevation model (DEM) and pre-fire vegetation structure metrics, which are essential for calculating fuel consumption and topographic influence on fire spread post-burn. Many drones (including DJI drones) have built-in functionality for oblique mapping. If unavailable, third-party photogrammetry software such as DroneDeploy was used to generate mapping flight routines. Depending on site access and burn size, full coverage of the burn unit was not always feasible; in such cases, priority was given to regions expected to exhibit complex fire behavior or containing nadir thermal plots. Table \ref{tab:photogrammetry_settings} lists representative photogrammetry parameters used during data collection.

\begin{table}[htbp]
\renewcommand{\arraystretch}{1.3}
\centering
\caption{Pre- and Post-Burn Structure From Motion Mapping Settings}
\label{tab:photogrammetry_settings}
\begin{tabular}{c||c}
\hline
\bfseries Parameter & \bfseries Value\\
\hline\hline
Location & Center on Nadir Plot(s)\\
Altitude & 122 m AGL\\
Terrain Following & On \\
Oblique Angle & 50\degree \\
Side Overlap & 50\% \\
Front Overlap & 80\% \\
\hline
\end{tabular}
\vspace{-5pt}
\end{table}

Generic oblique imagery of the surrounding no-fire environment was collected to serve as negative samples for machine learning applications. For image pairs, the UAV was piloted to capture diverse viewpoints while minimizing repeat imagery. Videos were collected with similar goals, with additional emphasis on maintaining smooth footage. To ensure stability for photogrammetric reconstruction and visual analysis, gimbal pitch speed was restricted to $<10^{\circ}$/s and yaw smoothness values were maximized in the flight controller settings to prevent sudden jerks. Pixel-wise correspondence between RGB and thermal imagery can vary depending on UAV platform, gimbal motion, and aircraft dynamics. To reduce these effects, rapid UAV and gimbal accelerations were minimized whenever possible. Table \ref{oblique_imagery_settings} details suggested collection parameters.

\begin{table}[htbp]
\renewcommand{\arraystretch}{1.2}
\caption{Pre-Burn Oblique Imagery Collection Parameters}
\label{oblique_imagery_settings}
\centering
\begin{tabular}{c||c|c}
\hline
\bfseries Parameter & \bfseries Image Pairs & \bfseries Video Pairs\\
\hline\hline
Location & Pre-burn environment & Pre-burn environment\\
Altitude & 61 - 122 m AGL & 61 - 122 m AGL\\
Time Between Photos & 5 seconds & n/a \\
Video Durations & n/a & 4-5 minutes \\
\hline
\end{tabular}
\vspace{-5pt}
\end{table}

The last non-supplemental pre-burn objective is to set up and georeference reflective plates for nadir thermal plots. The plates act as ground control points for georeferencing all UAV-derived products, a known location and distance measurement, and help the pilot orient the UAV during imagery acquisition. Polished aluminum 'pizza pans' work notably well for our purposes as they contrast against the forest floor in the RGB light spectrum. Aluminum has a very low emissivity, so the pans also contrast well against fire in the infrared imagery. During the nadir plot data collection, we placed four plates in the center, north, east, and south directions, and high-resolution GPS points, preferably from an RTK (real-time kinematic) platform, were collected from the middle each plate. The asymmetrical shape help the pilot center the UAV directly over the nadir plot during fire imagery collection, and it is important to choose a plot location where the plates will be visible to the UAV. In dense canopy situations, additional plates were added, and an inner ``nested'' plot design on a smaller scale was utilized for orientation and fire behavior metrics.

Supplemental data were gathered to help further describe fuel conditions at the research sites. The use of portable weather stations with high sampling and recording rates were placed around the area to be burned for more site-specific temperature, relative humidity, and wind speed and direction data during the burn. Fuel moisture samples were collected throughout the identified thermal plot area, at varying heights within the fuel bed. The use of clip plots was utilized outside the plot area to get actual fuel loading data, and other supplemental non-destructive fuel loading data collection methodology, such as Photoload \cite{Photoload}, was implemented to help further describe the fuel bed within the area of focus.

\subsubsection{Active-Burn}
Our during-burn FLAME 3 objectives are as follows:
\begin{itemize}
    \item Generic oblique imagery
    \item Nadir thermal plots
    \item (Supplemental) Regular weather measurements
\end{itemize}

For optimal collection, a UAV was dedicated to nadir thermal plots while one or more other UAVs patrols the flaming front, capturing generic oblique imagery. For generic oblique imagery, the same collection parameters in Table \ref{oblique_imagery_settings} that were used for oblique imagery collection during pre-burn are suggested, with the exception that the imagery subject should be the flaming front or interesting burn behavior. For nadir thermal plots, the objective of the nadir plot imagery is to capture the fire behavior in an area representative of the entire fire. This sampling approach captures ignition, flaming combustion, and post-frontal consumption for each pixel (area) in the nadir plot as described in  \cite{Oneill}. The UAV is positioned 122 m (400 ft) directly over the center of the nadir plot and oriented north. Plots are established prior to ignitions with the four aluminum ground control points. Immediately prior to the fire entering the UAV sensor frame, we began sampling images on a 5-second interval (0.2 Hz), pausing occasionally to swap UAV batteries. Ideally, all fire radiative energy released is captured, though post-frontal smoldering can occur for hours and even days while coarse woody debris is consumed. We balanced capturing multiple nadir plots at each fire with measuring post-frontal energy release, sampling plots until radiant energy release had subsided.

\subsubsection{Post-Burn}

The post-burn portion of the FLAME~3 dataset includes post-fire structure-from-motion (SfM) photogrammetry. After burn completion, the same oblique mapping flight plan used during the pre-burn stage was repeated. This ensured direct spatial comparability between pre-burn and post-burn point clouds, enabling analysis of fire effects on vegetation structure and terrain.

\subsection{Data Processing}
This section assumes that data has been collected according to the previously outlined FLAME 3 imagery collection procedures. Example raw data can be found \href{https://www.kaggle.com/datasets/brycehopkins/flame-3-computer-vision-subset-sycan-marsh}{here.}

\subsubsection{Imagery}
There are two imagery data types collected during the burn: frame pairs and video pairs. The data processing pipeline we used for using raw thermal RJPEG data, converting it to radiometric thermal TIFFs, and generating the colormapped thermal JPEGs .

  

  

We include two software tools in the FLAME 3 dataset, found under \href{https://github.com/BryceHopkins14/Flame-Data-Pipeline}{FLAME Data Pipeline Tools:}
\begin{enumerate}
    \item Raw file sorting tool, which pairs RGB/Thermal images, extracts radiometric metadata and saves it as a TIFF, regenerates thermal JPGs using temperature values, aligns RGB and thermal image pairs, and sorts all paired images and videos to proper output folders.
    \item An image labeling tool, which allows for rapid manual Fire/No Fire labeling, corresponding output folder sorting, and temperature thresholding functionality.
\end{enumerate}

\begin{figure}[h]
  \centering
  \includegraphics[width=0.45\textwidth]{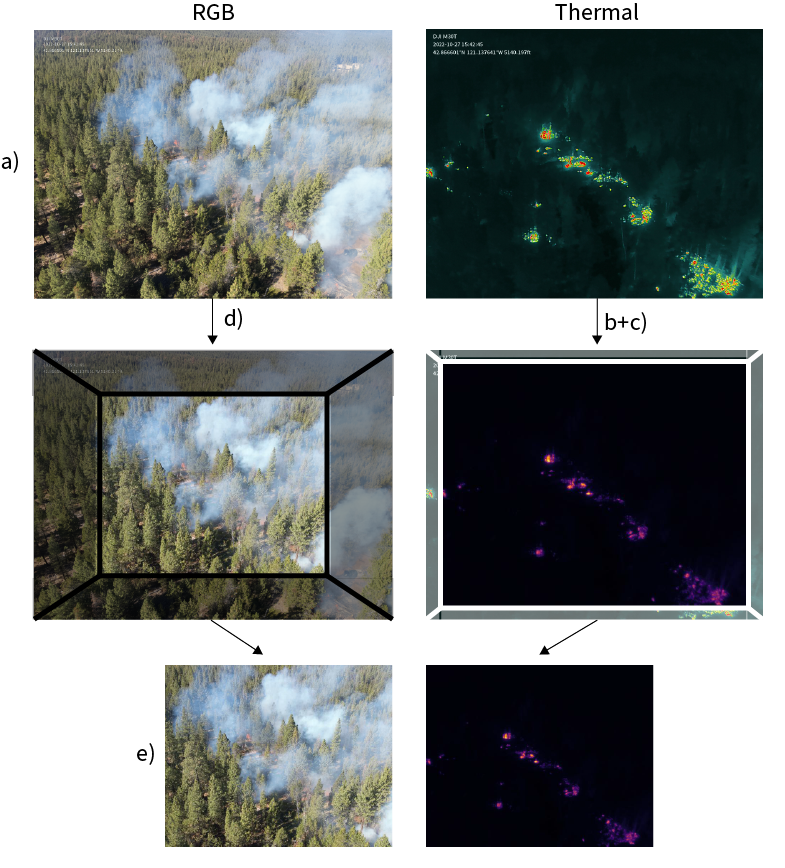}
  \caption{ FLAME 3 Imagery pre-processing workflow. Step (a) visualizes matching corresponding RGB and thermal images based on timestamp. Step (b) represents extracting out the radiometric thermal TIFF from the raw rJPEG and step (c) represents regeneration of the thermal image with the standardized "inferno" color palette. Step (d) represents the FOV corrections on the RGB image to match the dimensions of the thermal image. Step (e) represents the output of the paired RGB and Thermal Image.}
  
  \label{fig:ImageProcessFigure}
  \vspace{-5pt}
\end{figure}

\begin{figure}[h]
  \centering
  \includegraphics[width=0.45\textwidth]{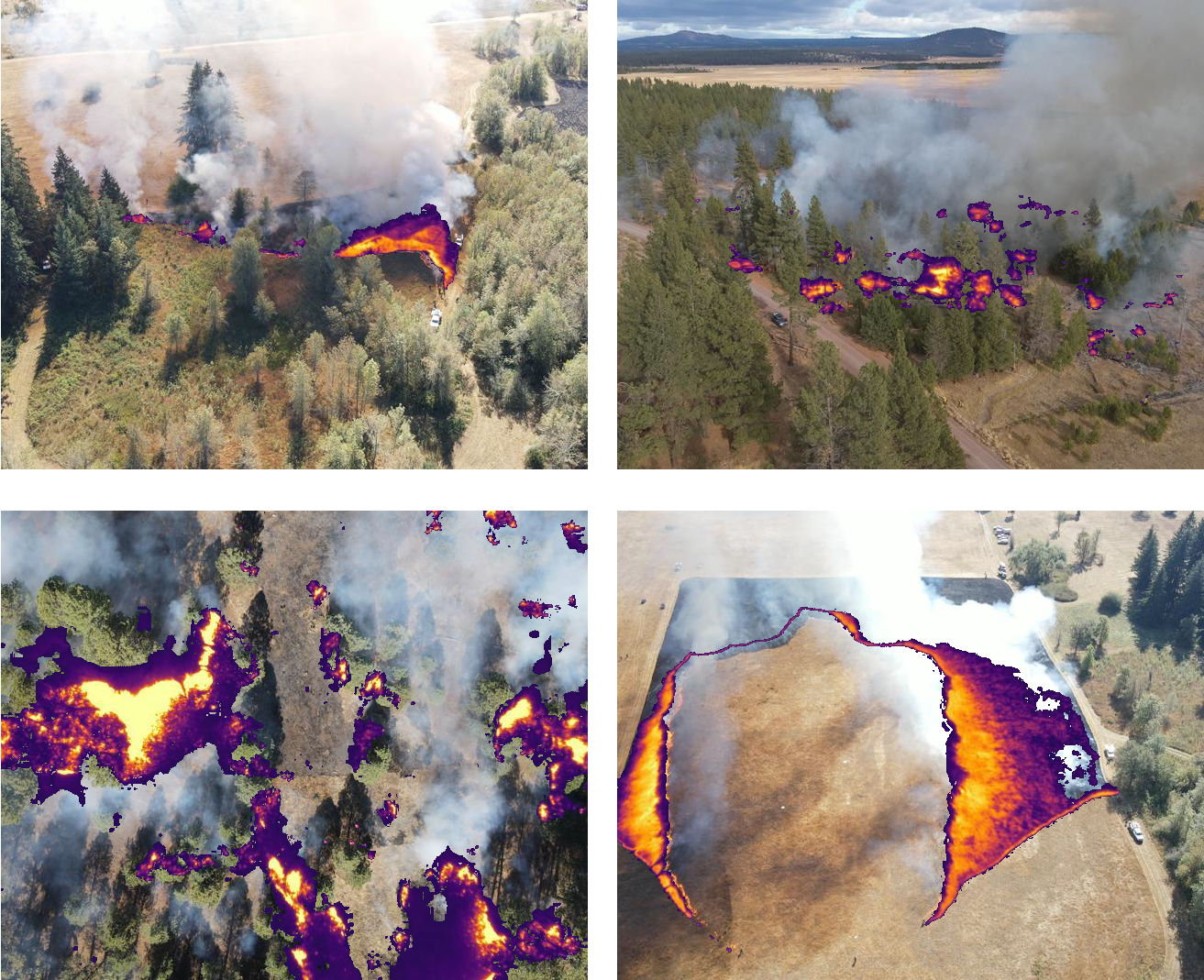}
  \caption{Overlaying of processed RGB-IR image pairs from Willamette (Top Left and Bottom Right), Sycan (Top Right), Shoetank (Bottom Left)}
  \label{fig:AlignmentFigure}
  \vspace{-5pt}
\end{figure}

\begin{figure}[h]
  \centering
  \includegraphics[width=0.45\textwidth]{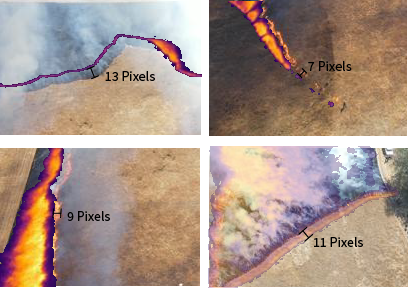}
  \caption{Error in RGB-IR image pair alignment associated with distortion from the IR camera in areas of the bottom right Willamette image from Figure 8 quantified using the Euclidean distance.}
  \label{fig:AlignmentErrorFigure}
  \vspace{-5pt}
\end{figure}

To address the inherent spatial discrepancies between the visible-spectrum and thermal sensors, spatial registration was performed to achieve pixel-level correspondence. Alignment was optimized iteratively by computing the necessary affine transformations (translation, scaling, and cropping) required to map the RGB coordinate space to the thermal sensor's field of view. The registration accuracy was validated by overlaying salient thermal features onto the transformed RGB images, as demonstrated in Figure~\ref{fig:AlignmentFigure}. To quantify the residual alignment error, pixel-wise displacement was measured using Euclidean distance across paired control points. As illustrated in Figure~\ref{fig:AlignmentErrorFigure}, maximum residual offsets were constrained to approximately 20 pixels, primarily driven by uncorrected radial lens distortion at the image extremities. Future dataset iterations will incorporate automated camera calibration matrices to mathematically correct this radial distortion and further minimize registration error.


\subsubsection{Photogrammetry}
Images sampled for pre- and post-fire point cloud construction were processed using Agisoft Metashape (version 2.0), a photogrammetric software package that applies structure-from-motion techniques to generate point clouds, orthomosaics, and related geospatial products \cite{Agisoft2023}. Processing parameters followed recommendations from Tinkham et al.~\cite{tink}. Point clouds were georeferenced using ground control points collected with RTK GPS.

While exhaustive independent check-point error reports were not systematically extracted for every individual flight, alignment observations and the inherent precision of the RTK-enabled UAV platforms (DJI M30T and M2EA) tied to the reflective aluminum ground control points yielded expected root-mean-square error (RMSE) values were expected to be on the order of 2--5~cm horizontally and 5--10~cm vertically. This level of precision indicates sufficient spatial accuracy for reliable plot-scale fire behavior analysis, thermal mapping, and vegetation structural assessment.

\section{FLAME Classification Comparison}
\label{sec:classification_results}

To evaluate the improvement of aerial imagery contained within FLAME~3 compared to previous FLAME~1 \cite{shamsoshoara2021aerial} and FLAME~2 \cite{FLAME2Dataset} iterations, models were trained for a binary Fire/No-Fire classification task. To directly address concerns of model over-parameterization and overfitting on smaller datasets, a custom lightweight Convolutional Neural Network (CNN) architecture was utilized in place of massive pre-trained models. The baseline model consists of three convolutional blocks (Conv2d, BatchNorm, ReLU, MaxPool2d) followed by a Global Average Pooling layer and a linear classifier. This constrained capacity forces the network to learn generalized fire features rather than memorizing training data. For multimodal inputs (paired RGB and Thermal imagery), we evaluated two distinct fusion strategies. The first is an ``Early Fusion'' baseline, where modalities are concatenated along the channel dimension prior to feature extraction. The second is a novel ``Cross-Attention Late Fusion'' architecture designed to intelligently leverage contextual cues. In this approach, independent CNN feature extractors process the RGB and Thermal streams. The resulting spatial feature maps are passed through a Multihead Cross-Attention mechanism, where the robust thermal signal acts as the query to dynamically extract relevant visual context (e.g., smoke density or flame color) from the RGB spatial keys and values before final classification.
To ensure transparency and address class imbalance, Table \ref{table:dataset_splits} details the specific training, validation, and testing splits. A rigorous, geographically isolated evaluation protocol was enforced: imagery from the Willamette and Sycan burn was strictly held out as the universal test set for all models. To prevent skewed evaluation metrics, the test set was randomly subsampled to ensure a perfectly balanced ratio of Fire to No-Fire images. 
During training, class imbalance was mitigated using a weighted random sampler to guarantee an equal distribution of classes within every batch. Hyperparameter optimization (learning rate, weight decay, base filter count, and dropout) was performed using Bayesian optimization via Optuna \cite{snoek_bayesian}. Models were evaluated on a static 20\% validation split during the hyperparameter sweep, followed by a final training run on the full training pool. In addition to accuracy, precision, and recall, we report the sample-independent Area Under the Receiver Operating Characteristic Curve (ROC-AUC) to provide a robust measure of model performance.

For single-modality experiments (RGB, Thermal JPG, and Thermal TIFF), the lightweight unimodal CNN was used directly. For paired multimodal experiments (RGB--Thermal JPG and RGB--TIFF), we evaluated both the early-fusion and cross-attention late-fusion variants described above. Single modality refers to using either RGB, Thermal, or TIFF data only as input to the model; multimodal refers to paired RGB--Thermal or RGB--TIFF inputs. Weights were initialized using Kaiming initialization \cite{he_delving} for convolutional layers and Xavier initialization \cite{glorot_xavier} for linear layers.

To ensure transparency regarding data distribution and to address potential concerns about class imbalance, Table \ref{table:dataset_splits} details the specific training, validation, and testing splits used for each experimental setup. While the training sets maintain the natural class imbalance observed during collection (often favoring Fire samples), the test sets were constructed to be balanced where possible or sufficiently large to provide statistically significant performance metrics.
The FLAME~1 classification set contained 30{,}155 Fire RGB images and 17{,}837 No-Fire RGB images. The FLAME~2 set contained 39{,}751 Fire RGB--T pairs and 13{,}700 No-Fire RGB--T pairs. The FLAME~3 set consisted of imagery from five burns: Willamette, Shoetank, Sycan Marsh, Payson Hundred Rx, and Hanna Hammock. Each evaluation scenario specifies the FLAME~3 training and testing subsets used.

To reduce the risk of data leakage and overly optimistic performance estimates, FLAME~3 splits were constructed to avoid overlap between training and testing imagery from the same burst-capture sequences whenever possible, and any explicitly sampled test images from a burn were removed from the corresponding training pool (details provided in the following subsections). Because the class distributions differ across datasets (e.g., Fire vs.\ No-Fire counts), we report multiple metrics (accuracy, precision, sensitivity/recall, specificity, and F1-score) to better characterize performance under potential class imbalance. All runs were performed using the same training procedure and evaluation protocol within each scenario.

Additionally, Bayesian optimization \cite{snoek_bayesian} was used to perform a hyperparameter search. This sweep was intended to identify reasonable hyperparameters for each input modality rather than to serve as an exhaustive tuning procedure. The optimization and loss setup for the reported results is summarized in Table \ref{tab:opt_setup_80epoch}. Each reported evaluation was trained for 80 epochs, with batch size determined by input modality and fusion strategy as listed in Table \ref{tab:opt_setup_80epoch}. The best observed validation and held-out test performance from these 80-epoch runs is reported. 


\begin{table}[htbp]
\caption{Dataset composition for the classification experiments. Training counts are shown for each setup. A separate strictly held-out test set, constructed from Willamette and Sycan, contains 430 balanced Fire/No-Fire image pairs and is used consistently for all reported evaluations. The validation set for each experiment is formed as 20\% of the corresponding training pool.}
\label{table:dataset_splits}
\setlength{\tabcolsep}{3pt}
\renewcommand{\arraystretch}{1.3}
\begin{center}
\begin{tabular}{l|cc}
\toprule
\textbf{Experiment} & \multicolumn{2}{c}{\textbf{Training Set}} \\
& \textbf{Fire} & \textbf{No-Fire} \\
\midrule
RGB (F1) & 30,155 & 17,837 \\
RGB (F2) & 39,751 & 13,700 \\
RGB (F3) & 3,710 & 2,023 \\
\midrule
RGB+Thermal (F2) & 39,751 & 13,700 \\
RGB+Thermal (F3) & 3,942 & 2,255 \\
RGB+TIFF (F3) & 3,942 & 2,255 \\
TIFF (F3) & 3,942 & 2,255 \\
\bottomrule
\end{tabular}
\end{center}
\footnotesize{Note: F1 = FLAME 1, F2 = FLAME 2, F3 = FLAME 3.}
\end{table}

\begin{table*}[t]
\caption{Optimization and Loss Setup Used for Reported 80-Epoch Results}
\label{tab:opt_setup_80epoch}
\centering
\setlength{\tabcolsep}{5pt}
\renewcommand{\arraystretch}{1.35}
\begin{tabular}{l l l l c c c }
\toprule
\textbf{Input Type} & \textbf{Fusion} & \textbf{Optimizer} & \textbf{Loss Function} & \textbf{LR} & \textbf{Weight Decay} & \textbf{Batch} \\
\midrule
RGB (all combos) & Early / N.A. & AdamW & CE (LS$=0.1$) $+ 0.15\,$SupCon & $5\times10^{-4}$ & $1\times10^{-4}$ & 64 \\
Thermal (JPG/TIFF) & Early / N.A. & AdamW & CE (LS$=0.05$) $+ 0.1\,$SupCon & $8\times10^{-4}$ & $5\times10^{-5}$ & 32 \\
RGBT (JPG/TIFF) & Early & AdamW & CE (LS$=0.1$) $+ 0.2\,$SupCon & $3\times10^{-4}$ & $1\times10^{-4}$ & 32 \\
RGBT (JPG/TIFF) & Late (Attn) & AdamW & CE (LS$=0.1$) $+ 0.1\,$SupCon & $1\times10^{-4}$ & $2\times10^{-3}$ & 16 \\
\bottomrule
\end{tabular}

\vspace{1mm}
\footnotesize
\textit{Notes:} (1) Class imbalance is handled by \texttt{WeightedRandomSampler} and class-weighted cross-entropy. (2) Reported 80-epoch runs utilize modality-specific tuning; base filters fixed at 16 and dropout at 0.2. (3) SupCon temperature $\tau$ was tuned to 0.07 for RGB/RGBT and 0.1 for Thermal-only streams. (4)Abbreviations: LR (Learning Rate), CE (Cross-Entropy), LS (Label Smoothing), SupCon (Supervised Contrastive Loss).
\end{table*}

\subsection{Unimodal Evaluation}

To evaluate the effect of FLAME~3 data collection and preprocessing improvements on unimodal classification performance, experiments were conducted under RGB-only, Thermal JPG-only, and Thermal TIFF-only input settings. The FLAME~3 training pool contained imagery from Shoetank, Payson Hundred Rx, and Hanna Hammock. A separate strictly held-out test set was constructed from Willamette and Sycan and contains 430 balanced Fire/No-Fire image pairs. This test set was excluded from the training pool to eliminate leakage and to provide a geographically distinct evaluation setting.

\begin{table*}[htbp]
\caption{Classification Results: Unimodal Comparison. Evaluating domain shift across dataset combinations and the performance of single-modality RGB and Thermal inputs on the strictly held-out Willamette/Sycan test set.}
\label{tab:class_uni}
\setlength{\tabcolsep}{3pt}
\renewcommand{\arraystretch}{1.3}
\begin{center}

\begin{tabular}{>{\centering\arraybackslash}m{0.14\linewidth} >{\centering\arraybackslash}m{0.14\linewidth} >{\centering\arraybackslash}m{0.08\linewidth} >{\centering\arraybackslash}m{0.08\linewidth} >{\centering\arraybackslash}m{0.09\linewidth} >{\centering\arraybackslash}m{0.09\linewidth} >{\centering\arraybackslash}m{0.09\linewidth} >{\centering\arraybackslash}m{0.09\linewidth} >{\centering\arraybackslash}m{0.08\linewidth}}
\toprule
\textbf{Train Set} & \textbf{Modality} & \textbf{Val. Acc (\%)} & \textbf{Test Acc (\%)} & \textbf{Precision (\%)} & \textbf{Sensitivity (\%)} & \textbf{Specificity (\%)} & \textbf{F1-Score (\%)} & \textbf{AUC} \\[0.2mm]
\midrule
FLAME 1 & RGB & 99.86 & 50.70 & 50.92 & 38.60 & 62.79 & 43.92 & 0.6090 \\
FLAME 2 & RGB & 100.00 & 45.58 & 31.37 & 7.44 & 83.72 & 12.03 & 0.4255 \\
FLAME 3 & RGB & 97.46 & 56.51 & 53.63 & 96.28 & 16.74 & 68.89 & 0.7185 \\
FLAME 1 \& 2 & RGB & 99.88 & 40.00 & 38.97 & 35.35 & 44.65 & 37.07 & 0.4204 \\
FLAME 1 \& 3 & RGB & 99.71 & 60.70 & 68.55 & 39.53 & 81.86 & 50.15 & 0.6498 \\
FLAME 2 \& 3 & RGB & 99.82 & 53.02 & 63.27 & 14.42 & 91.63 & 23.48 & 0.6855 \\
FLAME 1, 2, \& 3 & RGB & 99.74 & 54.19 & 61.54 & 22.33 & 86.05 & 32.76 & 0.6609 \\
\midrule
FLAME 2 & THERMAL (JPG) & 100.00 & 66.51 & 61.95 & 85.58 & 47.44 & 71.87 & 0.7326 \\
FLAME 3 & THERMAL (JPG) & 100.00 & 97.91 & 95.98 & \textbf{100.00} & 95.81 & 97.95 & \textbf{1.0000} \\
FLAME 3 & THERMAL (TIFF) & 100.00 & \textbf{98.37} & 96.85 & \textbf{100.00} & 96.74 & \textbf{98.40} & \textbf{1.0000} \\
FLAME 2 \& 3 & THERMAL (JPG) & 100.00 & 93.26 & \textbf{100.00} & 86.51 & \textbf{100.00} & 92.77 & 0.9994 \\
\bottomrule
\end{tabular}
\end{center}
\end{table*}

As detailed in Table \ref{tab:class_uni}, all unimodal models achieved exceptionally high validation accuracy during training (often approaching 100\%). However, this metric must be interpreted carefully. Because the validation sets were generated via random splitting from the training pool—which consists of extracted continuous video frames—adjacent frames in the training and validation sets share nearly identical environmental backgrounds. Consequently, models can easily achieve high validation accuracy by memorizing domain-specific background features (e.g., specific soil colors or vegetation types) rather than learning generalized fire characteristics.

The true measure of generalization is the model's performance on the strictly held-out Willamette and Sycan test set. Here, the results directly address the relationship between dataset size and model robustness. Despite containing over 50,000 images, models trained exclusively on FLAME~2 suffered a complete generalization collapse when tested on the unseen geographic location, yielding a test accuracy of 45.58\% and an AUC of 0.4255. An AUC below 0.50 indicates severe domain shift; the model confidently predicted the wrong classes because it was relying on memorized FLAME~2 backgrounds that did not exist in the Willamette dataset.

Conversely, the inclusion of the much smaller FLAME~3 dataset drastically improved cross-geographic generalization across both visual and thermal modalities. In the visual spectrum, the model trained exclusively on FLAME~3 achieved the highest standalone RGB AUC (0.7185) and an exceptional sensitivity (96.28\%), proving its ability to robustly detect active fire across novel environments despite optical obscurants. However, the unimodal thermal models demonstrated absolute dominance over the RGB models. The model trained exclusively on FLAME~3 Thermal TIFFs achieved the highest overall unimodal performance, reaching 98.37\% test accuracy, a 98.40\% F1-Score, and a perfect 1.0000 AUC on the held-out test set. This mathematically confirms that radiometric TIFF values provide a powerful, physically grounded signal (absolute temperature) that allows the model to cleanly separate fire from background anomalies without being hindered by smoke, lighting variations, or the artifacts introduced by 8-bit color-mapped JPEGs.

\begin{figure*}[t]
  \centering

  \begin{subfigure}{0.5\textwidth}
    \centering
    \includegraphics[
      width=\linewidth
    ]{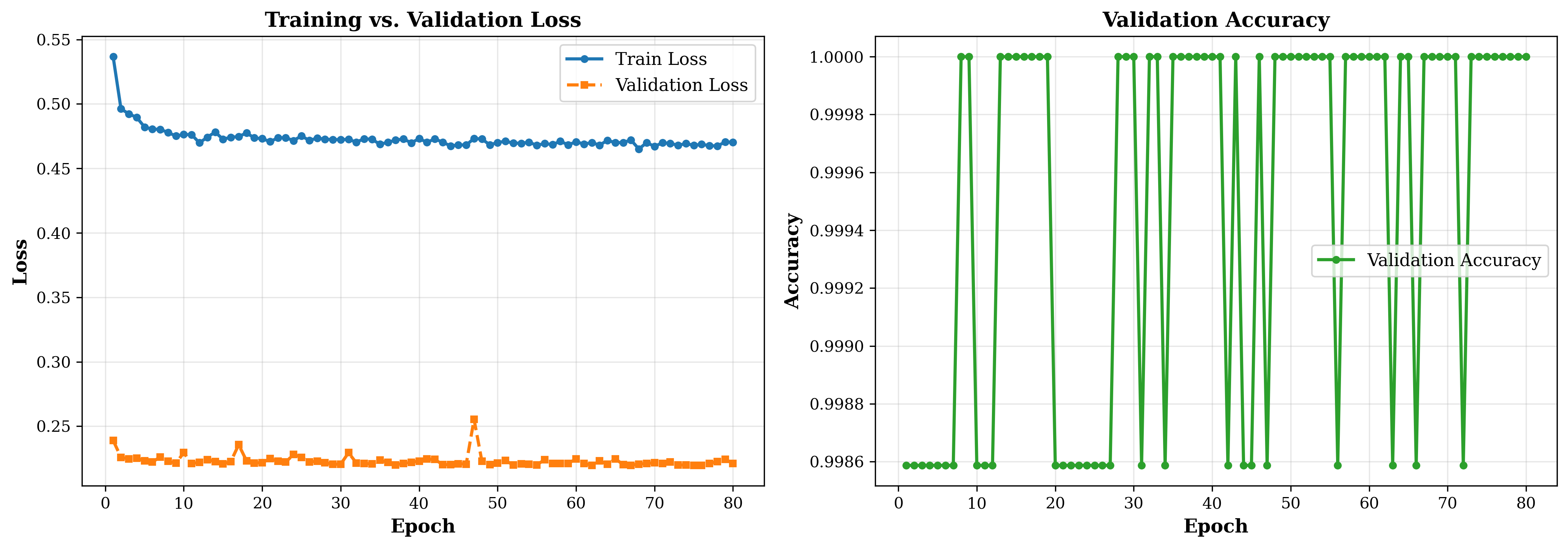}
    \caption{Training vs. Validation}
  \end{subfigure}
  \hfill
  \begin{subfigure}{0.2\textwidth}
    \centering
    \includegraphics[
      width=\linewidth
    ]{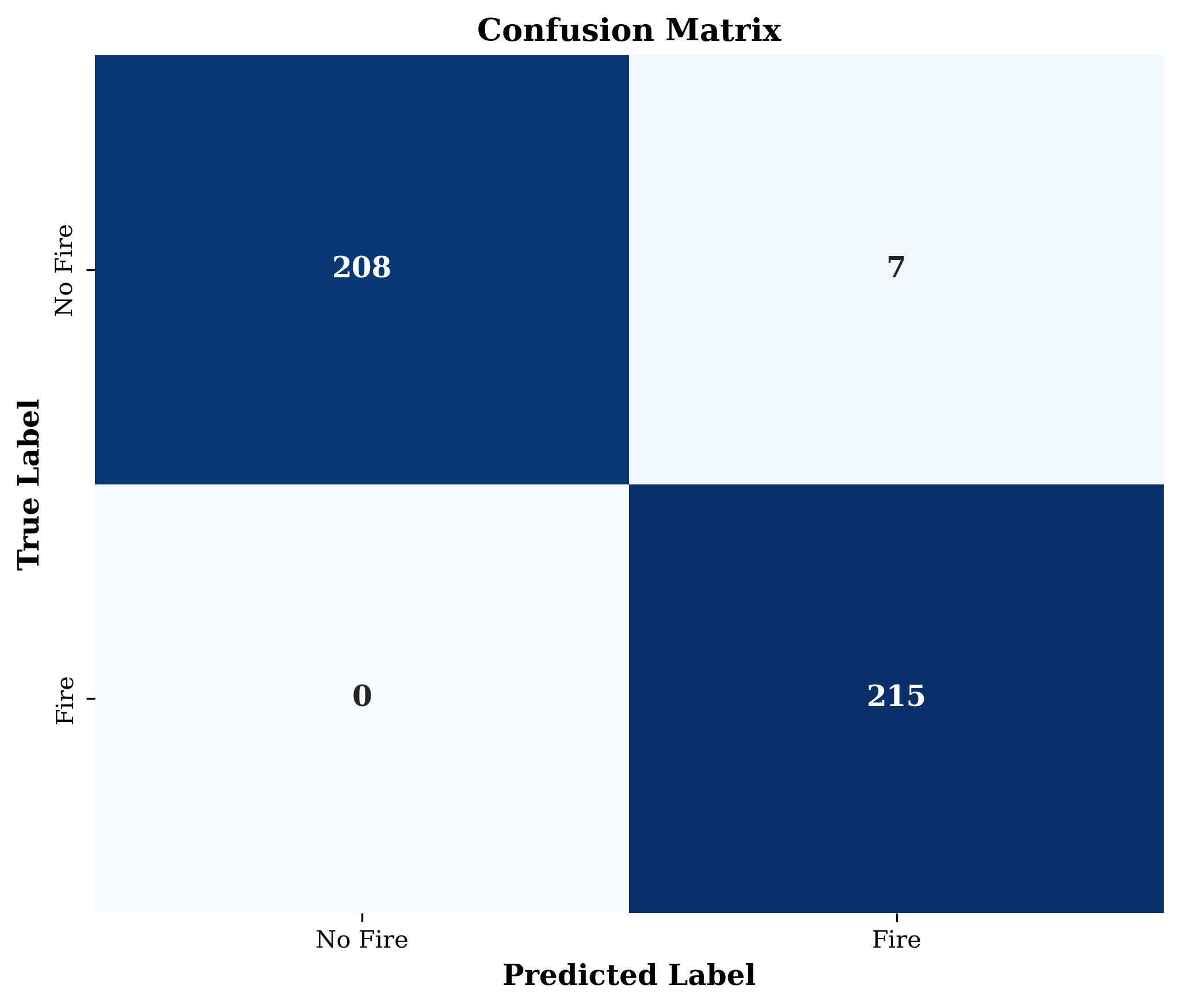}
    \caption{Confusion Matrix}
  \end{subfigure}
  \hfill
  \begin{subfigure}{0.2\textwidth}
    \centering
    \includegraphics[
      width=\linewidth
    ]{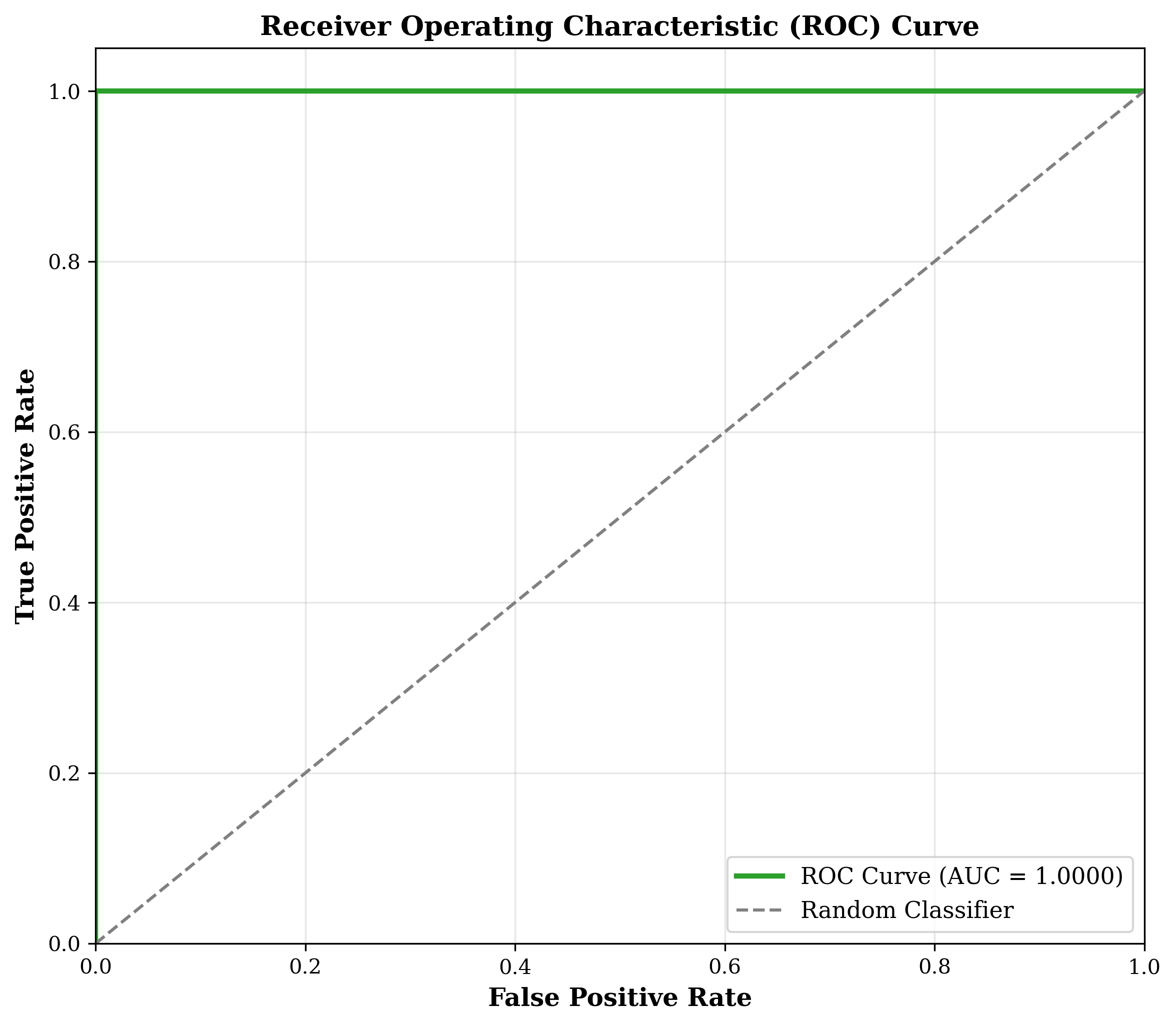}
    \caption{ROC Curve}
  \end{subfigure}

  \caption{Comprehensive evaluation metrics for the highest-performing unimodal architecture (FLAME~3 Thermal TIFF). By leveraging absolute radiometric temperature data, it achieved a perfect AUC (1.0000) and 98.37\% accuracy on the strictly held-out Willamette and Sycan test set.}
  \label{fig:best_model_metrics}
\end{figure*}

Figure~\ref{fig:best_model_metrics} illustrates the comprehensive evaluation metrics for the highest-performing unimodal architecture (FLAME~3 Thermal TIFF), including the training versus validation loss trajectories, the confusion matrix, and the ROC curve evaluated on the held-out test set. The stable convergence behavior of the loss curves indicates minimal overfitting, proving that the observed performance improvements are driven by generalized feature extraction rather than data leakage.

While Table \ref{tab:class_uni} summarizes the aggregated performance metrics, the complete set of learning curves (demonstrating the absence of overfitting), confusion matrices, and ROC curves for all unimodal configurations are provided in the Appendix.

\begin{table*}[htbp]
\caption{Classification Results: Multimodal Comparison. Evaluating early vs. late fusion architectures for paired visible/thermal inputs.}
\label{tab:class_multi}
\setlength{\tabcolsep}{3pt}
\renewcommand{\arraystretch}{1.3}
\begin{center}

\begin{tabular}{>{\centering\arraybackslash}m{0.14\linewidth} >{\centering\arraybackslash}m{0.15\linewidth} >{\centering\arraybackslash}m{0.08\linewidth} >{\centering\arraybackslash}m{0.08\linewidth} >{\centering\arraybackslash}m{0.09\linewidth} >{\centering\arraybackslash}m{0.10\linewidth} >{\centering\arraybackslash}m{0.10\linewidth} >{\centering\arraybackslash}m{0.09\linewidth} >{\centering\arraybackslash}m{0.08\linewidth}}
\toprule
\textbf{Train Set} & \textbf{Modality (Fusion)} & \textbf{Val. Acc (\%)} & \textbf{Test Acc (\%)} & \textbf{Precision (\%)} & \textbf{Sensitivity (\%)} & \textbf{Specificity (\%)} & \textbf{F1-Score (\%)} & \textbf{AUC} \\[0.2mm]
\midrule
FLAME 2 & RGBT JPG (Early) & 100.00 & 55.81 & 57.76 & 43.26 & 68.37 & 49.47 & 0.6723 \\
FLAME 2 & RGBT JPG (Late) & 100.00 & 62.79 & 64.40 & 57.21 & 68.37 & 60.59 & 0.6881 \\
\midrule
FLAME 3 & RGBT JPG (Early) & 100.00 & 93.95 & 89.54 & 99.53 & 88.37 & 94.27 & 0.9909 \\
FLAME 3 & RGBT JPG (Late) & 100.00 & \textbf{97.67} & \textbf{95.56} & \textbf{100.00} & \textbf{95.35} & \textbf{97.73} & \textbf{1.0000} \\
FLAME 3 & RGBT TIFF (Early)& 100.00 & 94.65 & 90.34 & \textbf{100.00} & 89.30 & 94.92 & 0.9945 \\
FLAME 3 & RGBT TIFF (Late) & 100.00 & 96.05 & 92.67 & \textbf{100.00} & 92.09 & 96.20 & \textbf{1.0000} \\
\midrule
FLAME 2 \& 3 & RGBT JPG (Early) & 100.00 & 86.74 & 88.35 & 84.65 & 88.84 & 86.46 & 0.8920 \\
FLAME 2 \& 3 & RGBT JPG (Late) & 100.00 & 86.05 & 80.63 & 94.88 & 77.21 & 87.18 & 0.9330 \\
\bottomrule
\end{tabular}
\end{center}
\footnotesize{Note: \textit{Test Acc} is calculated across 430 strictly held-out image pairs. \textit{Val. Acc} represents Optuna's internal training validation split performance.}
\end{table*}

\subsection{Multimodal Evaluation}
To evaluate the effectiveness of radiometric thermal TIFF data in conjunction with visual context, experiments were also conducted under multimodal input settings. FLAME~2 image counts were as described previously, and FLAME~1 was excluded from this evaluation because it does not provide paired thermal imagery designated for classification. Additionally, FLAME~2 does not include TIFF files. The FLAME~3 multimodal training pool contained imagery from Shoetank, Payson Hundred Rx, and Hanna Hammock, while the same strictly held-out test set constructed from Willamette and Sycan was used for evaluation. This test set contains 430 balanced Fire/No-Fire image pairs and was excluded from training to eliminate leakage and support rigorous cross-geographic testing. For RGB--TIFF combinations, Payson Hundred Rx and Hanna Hammock were removed where TIFF data were unavailable. TIFF raw data were normalized between 0 and 255 before conversion to PyTorch tensors. To ensure comparability across input types, the same preprocessing conventions and 80-epoch training schedule were used within each sweep.

The results from this testing are seen in Table \ref{tab:class_multi}. Among the evaluated multimodal networks, the novel FLAME~3 RGB--Thermal JPG Cross-Attention Late Fusion architecture achieved the highest overall test accuracy (97.67\%), an F1-Score of 97.73\%, and a perfect AUC of 1.0000. These configurations vastly outperform early-fusion multimodal inputs utilizing the thermal imagery from FLAME~2. Because the Willamette and Sycan test split is geographically isolated from the training data, these results firmly demonstrate robust cross-burn generalization without compromising sensitivity.

\begin{figure*}[t]
  \centering

  \begin{subfigure}{0.5\textwidth}
    \centering
    \includegraphics[
      width=\linewidth
    ]{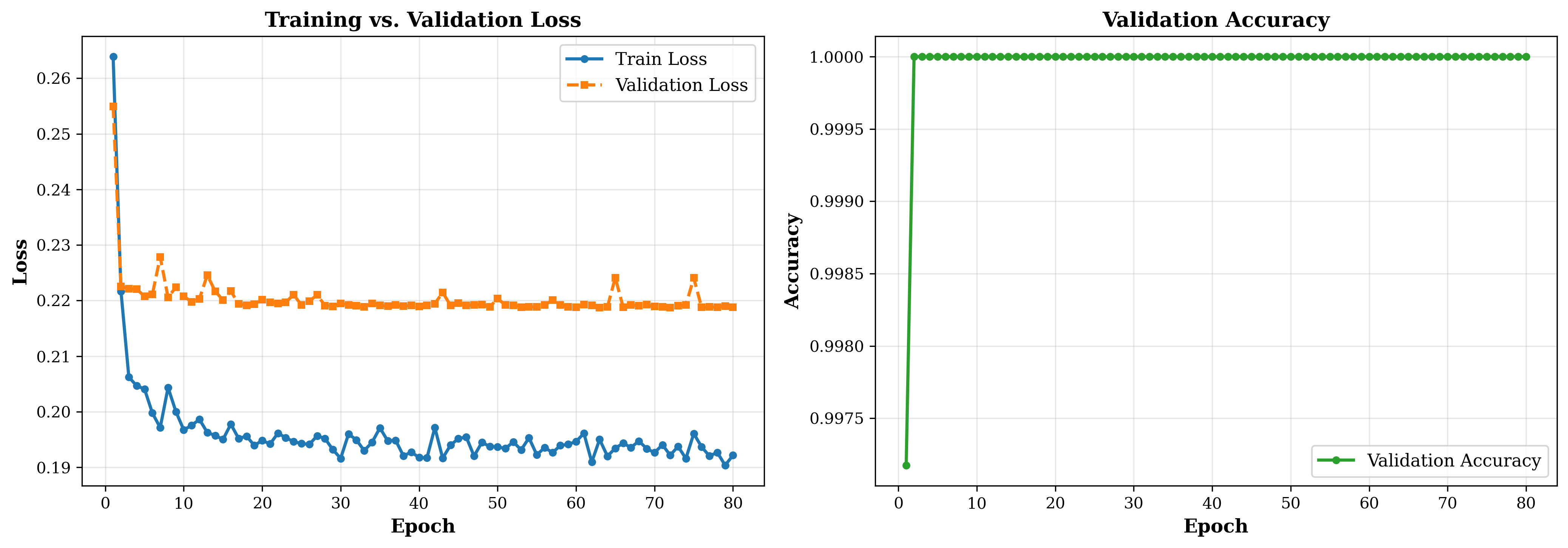}
    \caption{Training vs. Validation}
  \end{subfigure}
  \hfill
  \begin{subfigure}{0.2\textwidth}
    \centering
    \includegraphics[
      width=\linewidth
    ]{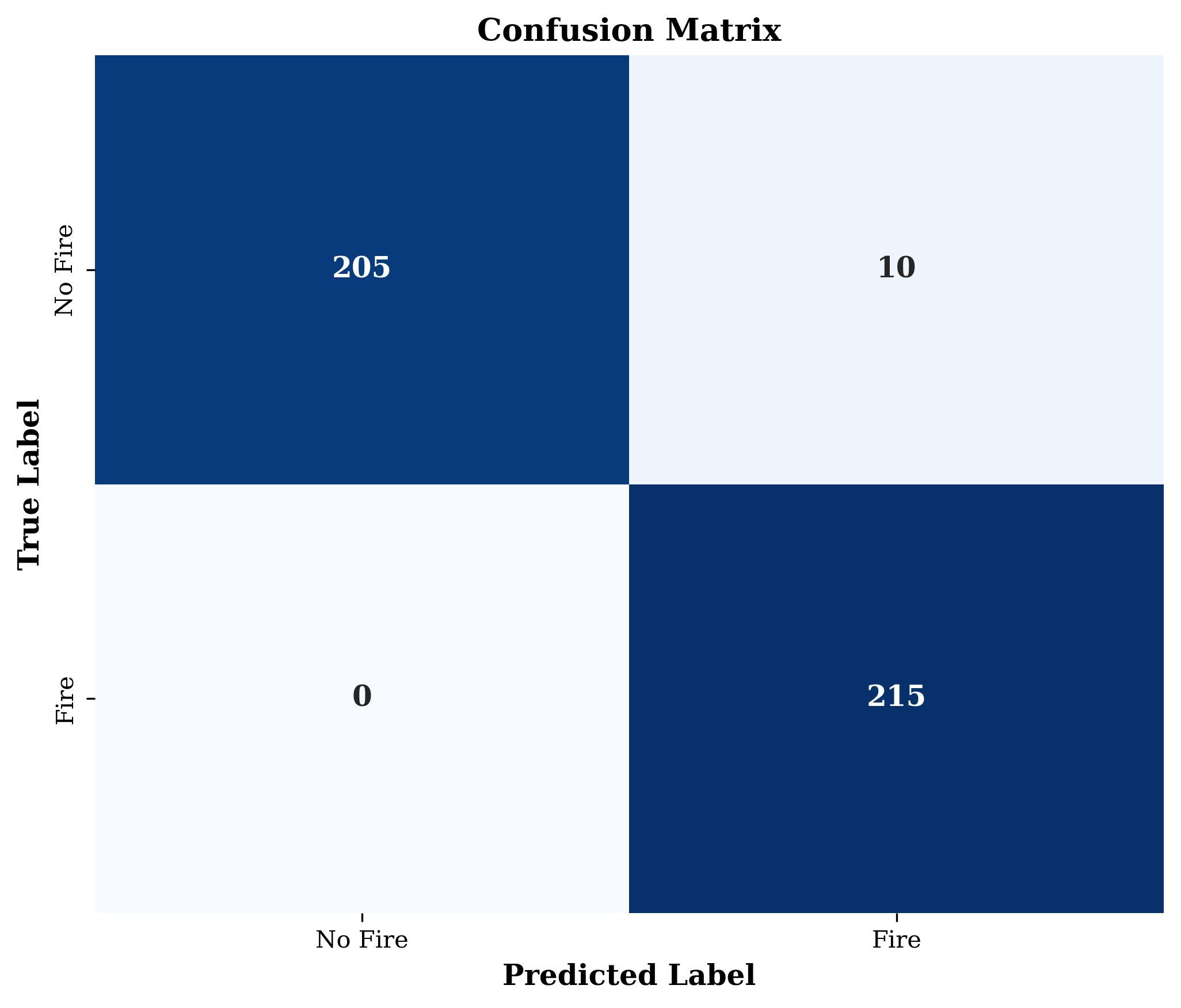}
    \caption{Confusion Matrix}
  \end{subfigure}
  \hfill
  \begin{subfigure}{0.2\textwidth}
    \centering
    \includegraphics[
      width=\linewidth
    ]{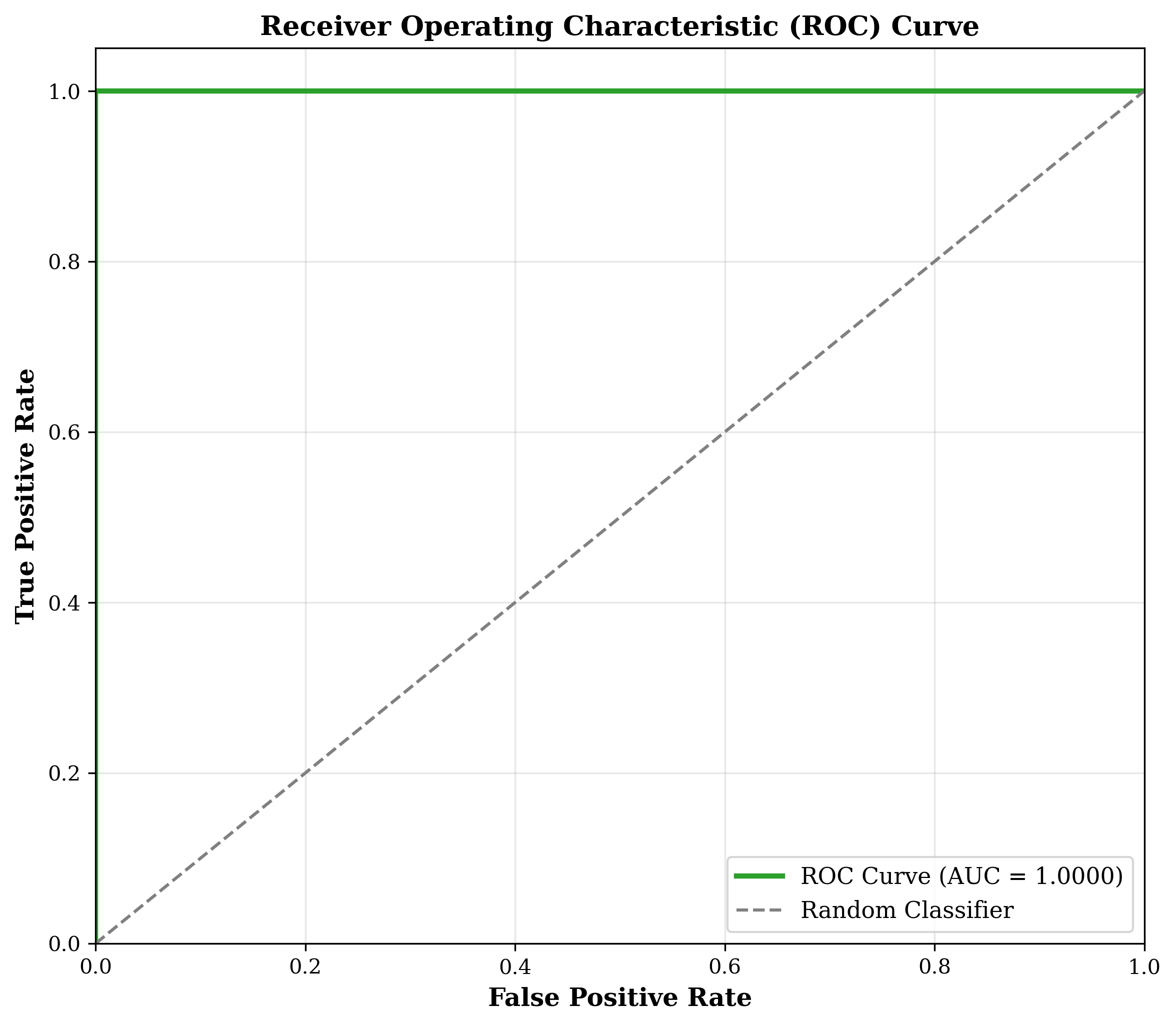}
    \caption{ROC Curve}
  \end{subfigure}

  \caption{Comprehensive evaluation metrics for the highest-performing FLAME~3 multimodal architecture (RGB-Thermal JPG Late Fusion), evaluated on the strictly held-out Willamette and Sycan test set.}
  \label{fig:best_multimodal_metrics}
\end{figure*}

Figure~\ref{fig:best_multimodal_metrics} illustrates the comprehensive evaluation metrics for the highest-performing multimodal architecture (FLAME~3 RGB-Thermal JPG Late Fusion). The close alignment of the loss trajectories indicates stable convergence with minimal overfitting, proving that the model successfully extracted generalizable multimodal features rather than memorizing the training distribution.

While RGB imagery contains valuable contextual cues (e.g., smoke plume structure and light scattering), these features can occasionally contradict the thermal signal. When evaluating simple early fusion architectures (feature concatenation), the model was easily misled by these conflicting signals; the dense optical noise from smoke degraded the thermal gradients, causing the FLAME~3 RGBT (Early) models to experience a performance drop.

However, we demonstrate that a robust network can leverage these RGB cues rather than be misled by them through the implementation of our Cross-Attention Late Fusion architecture. By isolating feature extraction and allowing the robust thermal features to actively query the RGB spatial map, the model successfully synthesized contextual smoke cues without compromising the primary thermal signal. Interestingly, while the unimodal evaluations in the previous section favored raw radiometric TIFFs, the multimodal cross-attention mechanism performed best with 8-bit colormapped Thermal JPEGs. This suggests that the high-contrast, non-linear color mappings of the JPEGs produce sharper spatial boundaries, making it easier for the attention mechanism to lock onto corresponding spatial structures in the RGB feature maps. This optimized spatial alignment effectively reduced false positives from 17 to 10 compared to the TIFF fusion variant, achieving a perfect 1.0000 AUC and proving that rigorous multimodal wildfire sensing requires dynamic spatial attention mechanisms.

Another important point is that the multimodal FLAME~3 inputs perform better than multimodal FLAME~2 inputs. For instance, the FLAME~2 RGBT (Early) model suffered severe domain shift, scoring an AUC of only 0.6723. The superior performance of the FLAME~3 multimodal models is consistent with FLAME~3's improved preprocessing (including precise FOV correction) and the higher spatial resolution of its thermal imagery.

Detailed visual diagnostics for all evaluated multimodal architectures, including epoch-by-epoch training and validation loss trajectories, precision-recall stability curves, and full confusion matrices, are exhaustively detailed in the Appendix.

\section{Conclusions} 
\label{sec:conclusions}

In this study, we introduced the FLAME~3 dataset to address a key data gap in AI-driven aerial wildfire monitoring. Unlike most existing wildfire image datasets, FLAME~3 provides synchronized visible-spectrum imagery and radiometric thermal TIFF files with quantitative per-pixel temperature measurements. Collected using UAVs at six prescribed fires, the dataset supports both generalized fire detection through oblique imagery and fine-grained temporal analysis through nadir thermal plots.

Our classification experiments highlight the practical value of these radiometric measurements. Models trained on FLAME~3 thermal TIFF inputs achieved 98.37\% test accuracy and 100\% sensitivity on the strictly held-out Willamette-and-Sycan test set, indicating that radiometric temperature data provides a strong and physically meaningful signal for distinguishing fire from non-fire conditions. More broadly, the unimodal and multimodal results suggest that FLAME~3 improves cross-burn generalization relative to prior FLAME dataset iterations and offers a useful benchmark for studying modality choice and fusion design.

Several limitations remain. Because FLAME~3 was collected during prescribed fires, it does not capture the full range of extreme fire behavior observed in large uncontrolled wildfires. In addition, residual lens-distortion-related alignment error between RGB and thermal imagery may affect dense pixel-level multimodal tasks. Accordingly, the reported classification study should be interpreted as a baseline evaluation demonstrating dataset utility rather than a definitive comparison of modeling strategies.

By releasing the dataset, collection methodology, and processing tools, we aim to broaden access to radiometric thermal wildfire imagery and support future work in detection, segmentation, temperature-aware modeling, and fire behavior analysis. The FLAME~3 dataset is publicly available through IEEE Dataport at
\href{https://ieee-dataport.org/open-access/flame-3-radiometric-thermal-uav-imagery-wildfire-management}{https://ieee-dataport.org/open-access/flame-3-radiometric-thermal-uav-imagery-wildfire-management}. Future work will focus on improving multimodal alignment, expanding dataset diversity, and developing fusion architectures that better combine radiometric thermal and RGB information.


\bibliographystyle{IEEEtran}
\bibliography{refV2}

\end{document}